\documentclass{article}

\usepackage[final]{neurips_2022}

\usepackage{natbib}
\usepackage[utf8]{inputenc} % allow utf-8 input
\usepackage[T1]{fontenc}    % use 8-bit T1 fonts
\usepackage{hyperref}       % hyperlinks
\usepackage{url}            % simple URL typesetting
\usepackage{booktabs}       % professional-quality tables
\usepackage{amsfonts}       % blackboard math symbols
\usepackage{nicefrac}       % compact symbols for 1/2, etc.
\usepackage{microtype}      % microtypography
\usepackage{graphicx}
\usepackage{sidecap}
\usepackage{wrapfig}
\usepackage{xcolor}
\usepackage{multicol}

\newcommand*\samethanks[1][\value{footnote}]{\footnotemark[#1]}

\usepackage{amsmath}
\usepackage{dsfont}
\newcommand{\mathbold}[1]{\ensuremath{\boldsymbol{\mathbf{#1}}}}

\usepackage[algoruled,noend]{algorithm2e}
\usepackage{listings}
\usepackage{fancyvrb}
\fvset{fontsize=\small}

\DeclareMathOperator{\EX}{\mathbb{E}}% expected value

\DeclareMathOperator*{\argmin}{arg\,min}

\newcommand{\bx}{\mathbold{x}}

\newcommand{\bkappa}{\mathbold{\kappa}}

\newcommand{\bphi}{\mathbold{\phi}}

\newcommand{\btau}{\mathbold{\tau}}
\newcommand{\btheta}{\mathbold{\theta}}

\title{Truncated proposals for scalable and hassle-free \\simulation-based inference}
\author{%
  Michael Deistler \\
  University of T\"{u}bingen\\
  \texttt{michael.deistler@uni-tuebingen.de} \\
  \And
  Pedro J Gon\c{c}alves\thanks{Equal contribution} \\
  University of T\"{u}bingen\\
  \texttt{pedro.goncalves@uni-tuebingen.de} \\
  \And
  Jakob H Macke\samethanks \\
  University of T\"{u}bingen\\
  Max Planck Institute for Intelligent Systems\\
  \texttt{jakob.macke@uni-tuebingen.de} \\
}

\begin{document}

\maketitle

\begin{abstract}
Simulation-based inference (SBI) solves statistical inverse problems by repeatedly running a stochastic simulator and inferring posterior distributions from model-simulations. To improve simulation efficiency, several inference methods take a sequential approach and iteratively adapt the proposal distributions from which model simulations are generated. However, many of these sequential methods are difficult to use in practice, both because the resulting optimisation problems can be challenging and efficient diagnostic tools are lacking. To overcome these issues, we present Truncated Sequential Neural Posterior Estimation (TSNPE). TSNPE performs sequential inference with truncated proposals, sidestepping the optimisation issues of alternative approaches. In addition, TSNPE allows to efficiently perform coverage tests that can scale to complex models with many parameters. We demonstrate that TSNPE performs on par with previous methods on established benchmark tasks. We then apply TSNPE to two challenging problems from neuroscience and show that TSNPE can successfully obtain the posterior distributions, whereas previous methods fail. Overall, our results demonstrate that TSNPE is an efficient, accurate, and robust inference method that can scale to challenging scientific models.
\end{abstract}

\section{Introduction}
\label{sec:introduction}
% Bayesian inference in mechanistic models.
Computational models are an important tool to understand physical processes underlying empirically observed phenomena. These models, often implemented as numerical  \emph{simulators}, incorporate mechanistic knowledge about the physical process underlying data generation, and thereby provide an interpretable model of empirical observations. In many cases, several parameters of the simulator have to be inferred from data, e.g., with Bayesian inference. However, performing Bayesian inference in these models can be difficult: Running the simulator may be computationally expensive, evaluating the likelihood-function might be computationally infeasible, and the model might not be differentiable. In order to overcome these limitations, Approximate Bayesian Computation (ABC) methods \citep{beaumont2002approximate,BeaumontCornuet_09}, synthetic likelihood approaches \citep{wood2010statistical}, and neural network-based methods \citep[e.g.,][]{papamakarios2016fast, hermans2020likelihood, thomas2022likelihood} have been developed.

% Why is NPE great?
%
\begin{wrapfigure}{ht!}{0.23\textwidth}
\vspace{-3pt}
\ifdefined\NOFIGS
    [FIGURE SKIPPED]
    [Comment out line 3 of main.tex to include]
\else
    \includegraphics[width=\linewidth]{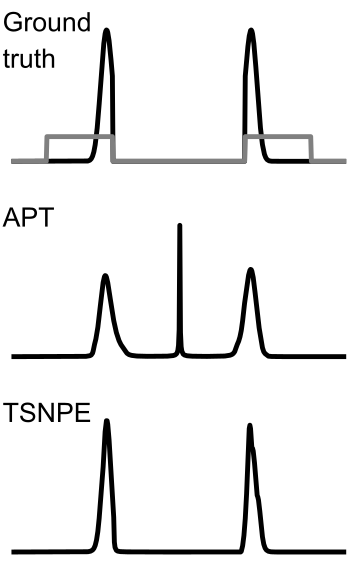}
\fi
\caption{
{\bf APT vs TSNPE.}
Top: Prior (gray) and true posterior (black). APT matches true posterior within the prior bounds but `leaks' into region without prior support. TSNPE (ours) matches true posterior.
}
\vspace{-10pt}
\label{fig:fig6}
\end{wrapfigure}
A subset of neural network-based methods, known as neural posterior estimation (NPE) \citep{papamakarios2016fast, lueckmann2017flexible, greenberg2019automatic}, train a neural density estimator on simulated data such that the density estimator directly approximates the posterior. Unlike other methods, NPE does not require any further Markov-chain Monte-Carlo (MCMC) or variational inference (VI) steps. As it provides an \emph{amortized} approximation of the posterior, which can be used to quickly evaluate and sample the approximate posterior for any observation, NPE allows the application in time-critical and high-throughput inference scenarios 
\citep{gonccalves2020training, radev2020bayesflow, dax2021real}, and fast application of diagnostic methods which require posterior samples for many different observations \citep{cook2006validation, talts2018validating}. In addition, unlike methods targeting the likelihood (e.g., neural likelihood estimation, NLE \citep{papamakarios2019sequential, lueckmann2019likelihood}), NPE can learn summary statistics from data and it can use equivariances in the simulations to improve the quality of inference \citep{dax2021real, dax2022group}.

% What is SNPE and why is it cool?
If inference is performed for a particular observation $\bx_o$, sampling efficiency of NPE can be improved with \emph{sequential} training schemes: Instead of drawing parameters from the prior distribution, they are drawn adaptively from a proposal (e.g., a posterior estimate obtained with NPE) in order to optimize the posterior accuracy for a particular $\bx_o$.
These procedures are called Sequential Neural Posterior Estimation (SNPE) \citep{papamakarios2016fast, lueckmann2017flexible, greenberg2019automatic} and have been reported to be more simulation-efficient than training the neural network only on parameters sampled from the prior, across a set of benchmark tasks \citep{lueckmann2021benchmarking}.

% Practical problems of SNPE.
Despite the potential to improve simulation-efficiency, two limitations have impeded a more widespread adoption of SNPE by practitioners: First, the sequential scheme of SNPE can be unstable. SNPE requires a modification of the loss function compared to NPE, which suffers from issues that can limit its effectiveness on (or even prevent their application to) complex problems (see Sec.~\ref{sec:background}). Second, several commonly used diagnostic tools for SBI \citep{talts2018validating, miller2021truncated, hermans2021averting} rely on performing inference across multiple observations. In SNPE (in contrast to NPE), this requires generating new simulations and network retraining for each observation, which often prohibits the use of such diagnostic tools \citep{lueckmann2021benchmarking, hermans2021averting}.

% We solve these issues with TSNPE
Here, we introduce Truncated Sequential Neural Posterior Estimation (TSNPE) to overcome these limitations. TSNPE follows the SNPE formalism, but uses a proposal which is a \emph{truncated} version of the prior: TSNPE draws simulations from the prior, but rejects them \emph{before simulation} if they lie outside of the support of the approximate posterior. Thus, the proposal is (within its support) proportional to the prior, which allows us to train the neural network with maximum-likelihood in every round and, therefore, sidesteps the instabilities (and hence `hassle') of previous SNPE methods. Our use of truncated proposals is strongly inspired by \citet{BlumFrancois_10} and \citet{miller2020simulation, miller2021truncated}, who proposed truncated proposals respectively for regression-adjustment approaches in ABC and for neural ratio estimation (see Discussion). Unlike methods based on likelihood(-ratio)-estimation \citep{miller2021truncated, hermans2021averting}, TSNPE allows direct sampling and density evaluation of the approximate posterior, and thus permits computing expected coverage of the full posterior quickly (without MCMC) and at every iteration of the algorithm, thus allowing to diagnose failures of the method even for high-dimensional parameter spaces (we term this `simulation-based coverage calibration' (SBCC), given its close connection with simulation-based calibration, SBC, \citet{cook2006validation, talts2018validating}). 
%Unlike previous methods that use expected coverage as a diagnostic \citep{dalmasso2020conditional, miller2021truncated, hermans2021averting}, computing the TSNPE expected coverage does not require evaluating the approximate posterior on a grid, building histograms of samples or MCMC sampling. Therefore, TSNPE expected coverage can be used to diagnose failures of the method even for high-dimensional parameter spaces.

We show that TSNPE is as efficient as the SNPE method `Automatic Posterior Transformation' (APT, \citet{greenberg2019automatic}) on several established benchmark problems (Sec.~\ref{sec:results:benchmark}). We then demonstrate that for two challenging neuroscience problems, TSNPE---but not APT---can robustly identify the posterior distributions (Sec.~\ref{sec:results:l5pc}).

% Paper overview
% After an introduction to neural network-based simulation-based inference (SBI), we present our method, Truncated Sequential Neural Posterior estimation (TSNPE). In Sec., we evaluate TSNPE on several benchmark tasks. In Sec., we demonstrate that TSNPE is scalable and robust: We obtain the posterior distribution of a complex neuroscience model, while previous methods fail to correctly infer the posterior distribution.

\section{Background}
\label{sec:background}

\begin{figure}[t!]
\ifdefined\NOFIGS
    [FIGURE SKIPPED]
    [Comment out line 3 of main.tex to include]
\else
    \includegraphics[width=\linewidth]{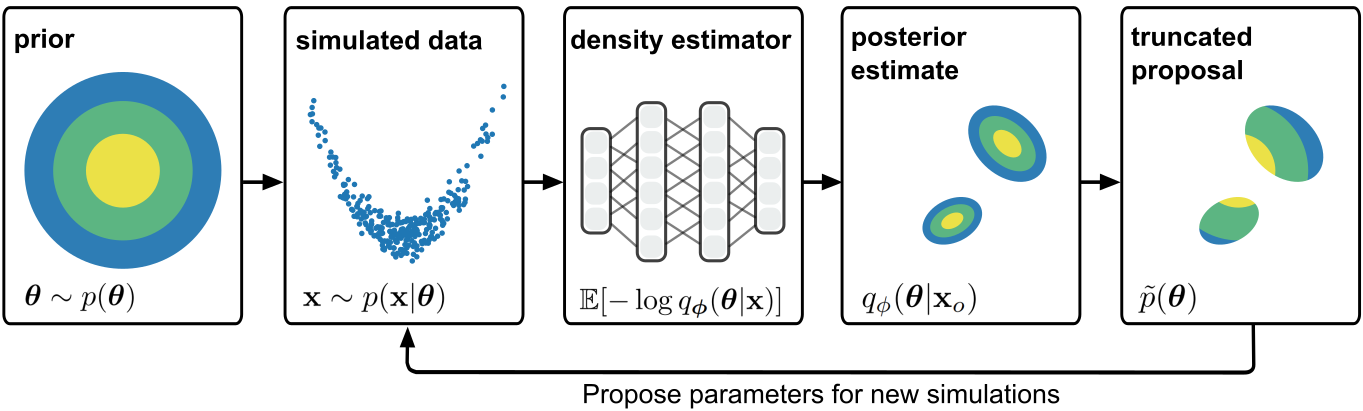}
\fi
\caption{
{\bf Truncated Sequential Neural Posterior Estimation (TSNPE).}
The method starts by sampling from the prior, running the simulator, and training a neural density estimator with maximum-likelihood to approximate the posterior. In subsequent rounds, parameters are sampled from the prior, but rejected if they lie outside of the support of the approximate posterior. With these proposals, the neural density estimator can be trained with maximum-likelihood in all rounds.
}

\label{fig:fig1}
\end{figure}

In Neural Posterior Estimation (NPE), parameters are sampled from the prior $p(\btheta)$ and simulated (i.e., $\bx$ is sampled from $p(\bx | \btheta)$). Then, a neural density estimator $q_{\bphi}(\btheta | \bx)$ (in our case a normalizing flow), with learnable parameters $\bphi$, is trained to minimize the loss:
\begin{equation*}
    \min_{\bphi} \mathcal{L} = \min_{\bphi}\EX_{\btheta \sim p(\btheta)}\EX_{\bx \sim p(\bx|\btheta)}[-\log q_{\bphi}(\btheta|\bx)],
\end{equation*}
which is minimized if and only if, for a sufficiently expressive density estimator, $q_{\bphi}(\btheta | \bx) = p(\btheta | \bx)$ for all $\bx \in \text{supp}(p(\bx))$ \citep{paige2016inference, papamakarios2016fast}. Throughout this study, we refer to training with this loss function as maximum-likelihood training, although the neural density estimator targets the posterior directly.

Sequential Neural Posterior Estimation (SNPE) aims to infer the posterior distribution $p(\btheta | \bx_o)$ for a particular observation $\bx_o$. SNPE initially performs NPE and, thereby, obtains an initial estimate of the posterior distribution. It then samples parameters from a proposal $\tilde{p}(\btheta)$, which is often chosen to be the previously obtained estimate of the posterior $\tilde{p}(\btheta) = q_{\bphi}(\btheta | \bx_o)$, and retrains the neural density estimator \citep{papamakarios2016fast}. This procedure can be repeated for several rounds. 

Importantly, if parameters $\btheta$ are sampled from the proposal $\tilde{p}(\btheta)$ rather than from the prior $p(\btheta)$, the estimator $q_{\bphi}(\btheta | \bx)$ that minimizes the maximum-likelihood loss function no longer converges to the true posterior. If one used the maximum-likelihood loss on data sampled from $\tilde{p}(\btheta)$, i.e., $\mathcal{L} = \EX_{\btheta \sim \tilde{p}(\btheta)}\EX_{\bx \sim p(\bx|\btheta)}[-\log q_{\bphi}(\btheta|\bx)]$, then $\mathcal{L}$ would be minimized by $q_{\bphi}(\btheta | \bx) \propto p(\btheta | \bx) \frac{\tilde{p}(\btheta)}{p(\btheta)}$, which is not the true posterior. Multiple schemes have been developed to overcome this \citep{papamakarios2016fast, lueckmann2017flexible}. 
The most recent of these methods, Automatic Posterior Transformation (APT, or SNPE-C, in its atomic version) \citep{greenberg2019automatic,durkan2020contrastive} employs a loss that aims to classify the parameter set that generated a particular data point among other parameter sets (details in Appendix Sec.~\ref{sec:appendix:apt}).

While APT has been reported to significantly outperform previous methods, several studies have also described cases in which the approach exhibits performance issues: 
Both the original APT paper \citep{greenberg2019automatic}  and \citet{durkan2020contrastive} reported that APT can show `leakage' of posterior mass outside of bounded priors. We demonstrate this issue on a simple 1-dimensional simulator with bounded prior (Fig.~\ref{fig:fig6}, Appendix Fig.~\ref{fig:fig6:supp2}). The posterior estimated by APT is only required to match the true posterior density \emph{within the support of the prior} (details in Appendix Sec.~\ref{sec:appendix:apt}). Thus, after five rounds of APT, while the approximate posterior matches the true posterior within the bounds of the prior, a substantial fraction of posterior mass lies in regions with zero prior probability. In simple models, approximate posterior samples that lie outside of the prior bounds can be efficiently rejected. However, in models with high numbers of parameters, the rejection rate can become so large that drawing posterior samples which lie inside of the prior bounds is prohibitive. For example, \citet{gloeckler2022variational} reported a rejection rate of more than 99.9999\% in a model with 31 parameters, thus requiring approximately one minute to draw a single posterior sample from within the prior bounds.

We overcome these limitations by using `truncated' proposal distributions. This allows us to train with maximum-likelihood at every round, thereby sidestepping issues of previous SNPE methods.

\section{Methodology}
\label{sec:methodology}

\begin{algorithm}[t]
 \textbf{Inputs:} prior $p(\btheta)$, observation $\bx_o$, simulations per round $N$, number of rounds $R$, $\epsilon$ that defines the highest-probability region ($\text{HPR}_{\epsilon}$)
 
 \textbf{Outputs:} Approximate posterior $q_{\bphi}$.
 
 \textbf{Initialize:} Proposal $\Tilde{p}(\btheta) = p(\btheta)$, dataset $\mathcal{X} = \{ \}$
 
 \For{$r \in [1, ..., R]$}{
    \For{$i \in [1, ..., N]$}{
        $\btheta_i \sim \Tilde{p}(\btheta)$ \\
        simulate $\bx_i \sim p(\bx|\btheta_i)$\\
        add $(\btheta_i, \bx_i)$ to $\mathcal{X}$
    }
    
    $\bphi^*=\argmin_{\bphi} - \frac{1}{N} \sum_{(\btheta_i,\bx_i) \in \mathcal{X}} \log q_{\bphi}(\btheta_i|\bx_i)$\\
    
    Compute expected coverage($\tilde{p}(\btheta), q_{\bphi}$) \tcp*{see Alg.~\ref{alg:alg2}}

    $\tilde{p}(\btheta) \propto p(\btheta) \cdot \mathds{1}_{\btheta \in \text{HPR}_{\epsilon}}$ \tcp*{see Alg.~\ref{alg:alg3}}
 }
\caption{TSNPE}
\label{alg:alg1}
\end{algorithm}

\subsection{Truncated proposals for SNPE}
\label{sec:methodology:truncated}
Given a particular observation $\bx_o$, we suggest to restrict the proposals $\tilde{p}(\btheta)$ to be proportional to the prior $p(\btheta)$ at least in the $1-\epsilon$ highest-probability-region ($\text{HPR}_{\epsilon}$, the smallest region that contains $1-\epsilon$ of the mass) of $p(\btheta | \bx_o)$, i.e.
\begin{equation*}
    \tilde{p}(\btheta) \propto p(\btheta) \cdot \mathds{1}_{\btheta \in \mathcal{M}}
\end{equation*}
with $\text{HPR}_{\epsilon}(p(\btheta | \bx_o)) \subseteq \mathcal{M}$. Thus, $\tilde{p}(\btheta)$ is a `truncated' proposal. The key insight is that, when using such a proposal and $\epsilon=0$, one can train $q_{\bphi}(\btheta | \bx)$ with maximum likelihood:
\begin{equation*}
    \min_{\bphi} \mathcal{L} = \min_{\bphi}\EX_{\btheta \sim \tilde{p}(\btheta)}\EX_{\bx \sim p(\bx|\btheta)}[-\log q_{\bphi}(\btheta|\bx)],
\end{equation*}
and $q_{\bphi}(\btheta | \bx_o)$ will still converge to $p(\btheta | \bx_o)$ (Proof in Appendix Sec.~\ref{sec:appendix:proofconvergence}).

We estimate $\mathcal{M}$ as the $\text{HPR}_{\epsilon}$ of the approximate posterior $\mathcal{M} = \text{HPR}_{\epsilon}(q_{\bphi}(\btheta | \bx_o))$. Since the maximum-likelihood loss employed to train $q_{\bphi}(\btheta|\bx)$ is support-covering, the $\text{HPR}_{\epsilon}$ of $q_{\bphi}(\btheta|\bx_o)$ tends to cover the $\text{HPR}_{\epsilon}$ of $p(\btheta|\bx_o)$ \citep{bishop2006pattern}.

In order to obtain the $\text{HPR}_{\epsilon}$ of $q_{\bphi}(\btheta | \bx_o)$, we define a threshold $\btau$ on the approximate posterior density $q_{\bphi}(\btheta | \bx_o)$. To do so, we use a normalizing flow as $q_{\bphi}(\btheta | \bx)$, which allows for closed-form density evaluation and fast sampling. We then approximate the $\text{HPR}_{\epsilon}$ of $q_{\bphi}(\btheta | \bx_o)$ as
\begin{equation*}
    \text{HPR}_{\epsilon}(q_{\bphi}(\btheta | \bx_o)) \approx \mathds{1}_{q_{\bphi}(\btheta | \bx_o) > \btau}.
\end{equation*}
We chose $\btau$ as the $\epsilon$-quantile of approximate posterior densities of samples from $q_{\bphi}(\btheta | \bx_o)$, and evaluated TSNPE for $\epsilon = 10^{-3}$, $10^{-4}$, and $10^{-5}$. Values of $\epsilon > 0$ yield a proposal prior which has smaller support than the current estimate of the posterior, e.g., using $\epsilon=10^{-3}$ neglects 0.1\% of mass from the approximate-posterior support. Thus, this approach leads to errors in posterior estimation, e.g., to `under-covered' posteriors (Appendix Sec.~\ref{sec:appendix:epsilon_errors}). However, empirically, the error induced by this truncation is negligible, as we will demonstrate on several benchmark tasks. We note that TSNPE can be trained on data pooled from all rounds (Appendix Sec.~\ref{sec:appendix:proofconvergence}). TSNPE is summarized in Alg.~\ref{alg:alg1} (Fig.~\ref{fig:fig1}).

\subsection{Sampling from the truncated proposal}
\label{sec:methodology:sampling}
To generate training data for subsequent rounds, we have to draw samples from the truncated proposal $\tilde{p}(\btheta)$, and here we explored rejection sampling and sampling importance resampling (SIR) \citep{rubin1988using}. For rejection sampling, we sample the prior $\btheta \sim p(\btheta)$ and accept samples only if their probability under the approximate posterior $q_{\bphi}(\btheta | \bx)$ is above threshold $\btau$. 

This strategy samples from the truncated proposal exactly, but can fail if the rejection rate becomes too high. To deal with these situations, we used SIR. For each sample from the truncated proposal, SIR draws $K$ samples from the approximate posterior, computes weights $w_{i=1...K} = p(\btheta_i)\mathds{1}_{\btheta_i \in \mathcal{M}} / q_{\bphi}(\btheta_i | \bx)$, normalizes $w_i$ such that they sum to one, draws from a categorical distribution with weights $s \sim \text{Categorical}(w_i)$, and selects the posterior sample with index $s$. SIR requires a fixed sampling budget of $K$ posterior samples per sample from the truncated proposal and returns exact samples from the truncated proposal for $K \rightarrow \infty$. Too low values of $K$ lead to too narrow proposals and posterior approximations. When run for a number of rounds, this behaviour reinforces itself and can lead to divergence of TSNPE (Appendix Fig.~\ref{fig:fig4:supp6}). We, thus, chose a high value $K=1024$. In our experiments, we did not observe poor SIR performance, but we emphasise the importance of using tools to diagnose potential failures of TSNPE (see below) or SIR (e.g.~by inspecting the effective sample size, Appendix Sec.~\ref{sec:appendix:sampling_sir_accuracy}). When SIR fails, methods such as nested sampling, adaptive multi-level splitting, or sequential Monte-Carlo sampling could be viable alternatives \citep{skilling2004nested, cerou2007adaptive, doucet2001sequential}. We discuss computational costs of rejection sampling and SIR in Appendix Sec.~\ref{sec:appendix:sampling_cost}.

% Alternatively, one could draw samples from the truncated proposal with nested sampling or Markov-chain Monte-Carlo algorithms, but we did not explore this possibility \citep{skilling2004nested}.

% For box uniform priors, as commonly used in multicompartment models of single-neuron dynamics, one can improve the acceptance rate of this scheme by obtaining bounds of the marginals of $q(\btheta | \bx)$. We do so by taking the minimum and maximum of $N$ samples of $q(\btheta | \bx)$ in each dimension, and replacing the prior with a uniform distribution within these bounds.

\subsection{Coverage diagnostic}
\label{sec:methodology:coverage}

In order for the estimated posterior $q_{\bphi}(\btheta | \bx_o)$ to converge to $p(\btheta | \bx_o)$, TSNPE requires $\text{supp}(p(\btheta | \bx_o)) \subseteq \text{HPR}_{\epsilon}(q_{\bphi}(\btheta | \bx_o))$, i.e., the estimated posterior must be broader than the true posterior (proof in Appendix Sec.~\ref{sec:appendix:proofconvergence}). In order to diagnose whether the posterior is, on average, sufficiently broad, we perform expected coverage tests as proposed in  \citet{dalmasso2020conditional, miller2021truncated, hermans2021averting}. 
%To obtain the empirical expected coverage, we need to compute the credible region with the smallest volume such that
%\begin{equation}
%    \int_{\btheta} q_{\bphi}(\btheta | \bx) d\btheta = 1 - \alpha
%\end{equation}
% with confidence level $1 - \alpha$. If the posterior is well-calibrated, the true data-generating parameter $\btheta^*$ should lie within this confidence interval with probability $1-\alpha$. 

\begin{SCfigure}
\caption{
{\bf Diagnostic tool.}
(a) Parameter $\theta^*$ (green) lies within the 1-$\alpha$ confidence region (gray) of the estimated posterior.
(b) $\log(p(\theta^* | x))$ is above the 1-$\alpha$ quantile of posterior samples.
(c) 1-$\alpha$ versus empirical coverage, averaged over $\theta^*$.
}
\includegraphics[width=0.65\textwidth]{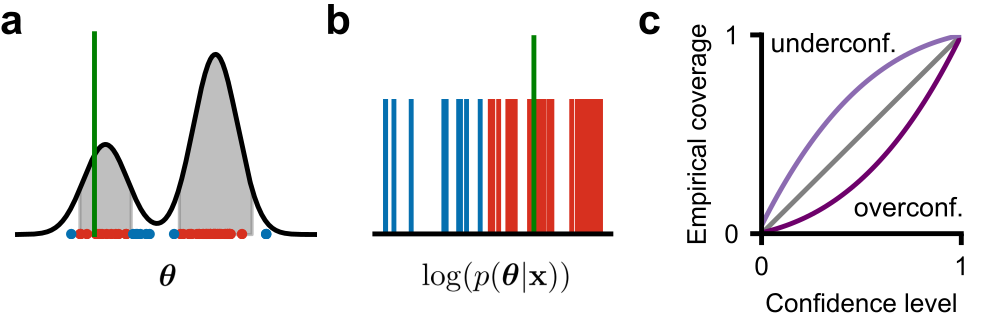}
\label{fig:fig2}
\end{SCfigure}

As described in \citet{dalmasso2020conditional, rozet2021arbitrary} and illustrated in Fig.~\ref{fig:fig2}, the coverage of the approximate posterior can be computed as
\begin{equation*}
    1-\alpha = \int q_{\phi}(\btheta | \bx^*) \mathds{1} (q_{\phi}(\btheta^* | \bx^*) \ge q_{\phi}(\btheta | \bx^*)) d\btheta
\end{equation*}
where $\btheta^*$ is sampled from the truncated proposal and $\bx^*$ is the corresponding simulator output. In order to approximate this integral, one has to either evaluate the approximate posterior on a grid \citep{dalmasso2020conditional, hermans2021averting} or apply a Monte-Carlo average which includes repeatedly sampling (and evaluating) the (unnormalized) approximate posterior \citep{miller2021truncated, rozet2021arbitrary}. The first option does not scale to high-dimensional spaces whereas the second is computationally expensive for methods estimating likelihood(-ratios) and, thus, require MCMC. In contrast, the TSNPE-posterior can be sampled from and evaluated in closed-form, leading to a computationally efficient and scalable diagnostic which can be run after every training round.

Expected coverage can be computed as an average of the coverage across multiple pairs $(\btheta^*, \bx^*)$ \citep{miller2021truncated, hermans2021averting} and should match the confidence level for all confidence levels $(1-\alpha) \in [0, 1]$ (Fig.~\ref{fig:fig2}c). We term this procedure of computing the empirical coverage `simulation-based coverage calibration` (SBCC), due to its close connection with SBC \citep{cook2006validation, talts2018validating} (identical under certain conditions, Appendix Sec.~\ref{sec:appendix:sbcrelation}). For TSNPE, it is important that the empirical expected coverage matches the confidence level for high confidence levels (i.e., for small $\alpha$), since overconfidence in these regions would indicate that ground-truth parameters $\btheta^*$ are falsely excluded from the $\text{HPR}_{\epsilon=\alpha}$. SBCC is summarized in Appendix Alg.~\ref{alg:alg2}.

\section{Results}
\label{sec:results}

We evaluated TSNPE on several benchmark tasks and on two complex problems from neuroscience. We found that TSNPE performs as well as APT on the benchmark tasks and that it is robust to choices of $\epsilon$. In addition, we found that, in contrast with APT, TSNPE can successfully infer the posterior distribution for complex models with large numbers of parameters.

% \subsection{Results on two moons task}

% \input{figures/fig3}
% To demonstrate how TSNPE infers the posterior distribution, we applied it the two-moons task
% \citep{greenberg2019automatic}. This model has two parameters with a uniform prior. The simulator is non-linear,
% generating a posterior with both local and global (bimodal) structure \citep{greenberg2019automatic,
% lueckmann2021benchmarking} (Fig 3a). We ran TSNPE for 10 rounds on this model with 10000 simulations in each round. The approximate posterior after the first round is shown in Fig3b, and its support in Fig3c. The support contains all out of 10000 samples from the true posterior (red dots in Fig 3c). A coverage test shows that that the posterior after the first round exhibits good coverage. Fig 3d shows the posterior after the final round, which closely matches the
% true posterior (classifier two-sample test accuracy: 0.53)
\begin{figure}[t]
\ifdefined\NOFIGS
    [FIGURE SKIPPED]
    [Comment out line 3 of main.tex to include]
\else
    \includegraphics[width=\linewidth]{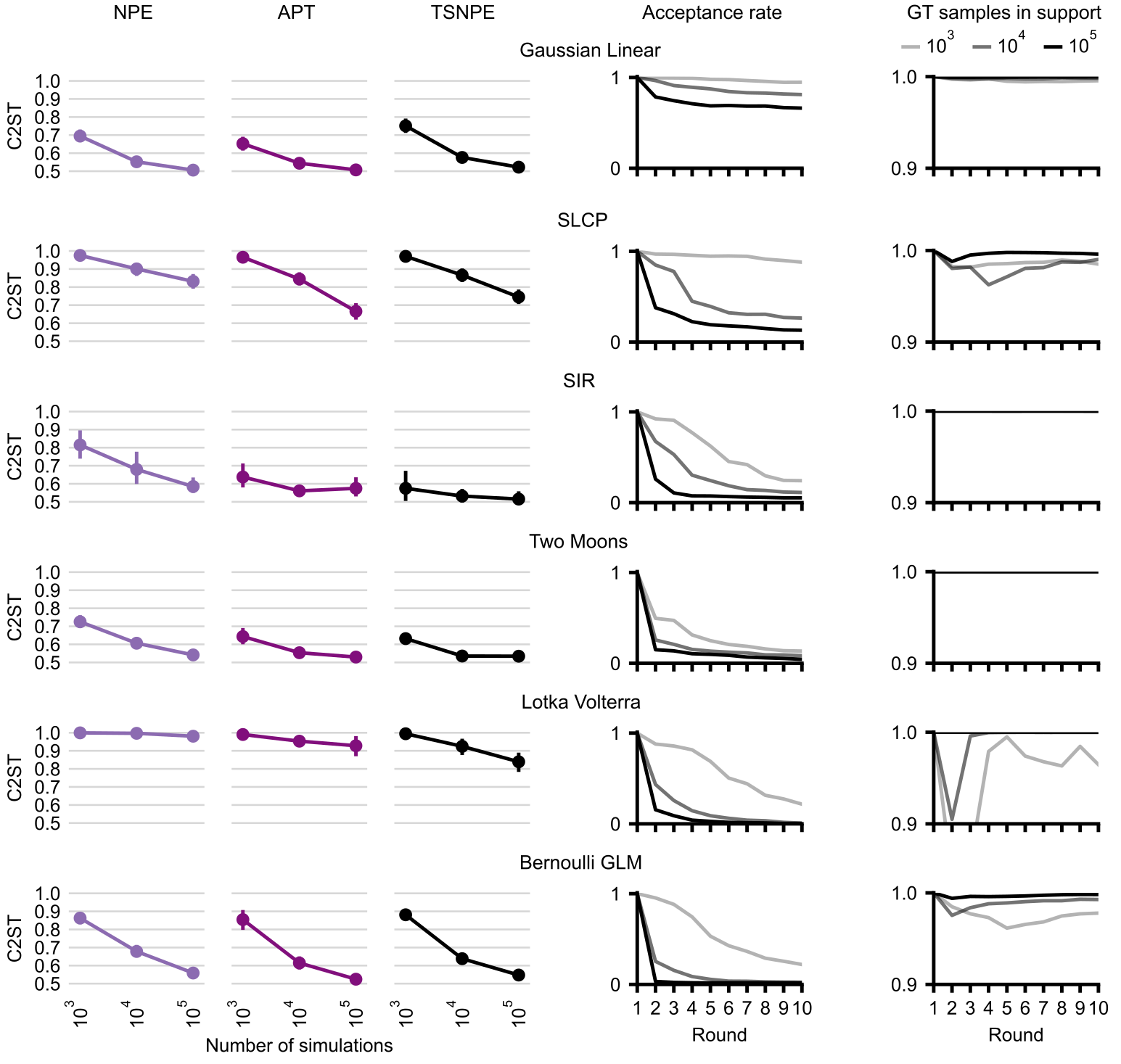}
\fi
\caption{
{\bf Performance on six benchmark tasks.}
Left three columns: classifier two-sample test accuracy (C2ST) of NPE (left), APT (middle), and TSNPE (right) for three simulation budgets. Forth column: Fraction of prior samples within the approximate-posterior $\text{HPR}_{\epsilon}$ in each round for each simulation budget. Fifth column: Fraction of true-posterior samples within the approximate-posterior $\text{HPR}_{\epsilon}$. TSNPE with $\epsilon = 10^{-4}$ and rejection sampling from truncated proposal.
}

\label{fig:fig4}
\end{figure}

\subsection{Performance on benchmark tasks}
\label{sec:results:benchmark}

We compared TSNPE with NPE and APT on six benchmark tasks for which samples from the ground-truth posterior are available (see Appendix Sec.~\ref{sec:appendix:benchmark} for tasks) \citep{lueckmann2021benchmarking}. We quantified the performance with a classifier two-sample test (C2ST), for which 0.5 indicates that the approximate posterior is identical to the ground-truth posterior, whereas 1.0 implies that the distributions can be completely separated by a classifier. Overall, APT and TSNPE perform similarly well and both outperform NPE (Fig.~\ref{fig:fig4}, left three columns). On two of the six tasks (Gaussian Linear and SLCP), APT has slightly better performance than TSNPE, whereas on two other tasks (SIR and Lotka-Volterra), TSNPE outperforms APT. Overall, TSNPE and APT perform similarly well, demonstrating that TSNPE is competitive with previous methods on benchmark tasks.

In order to get insights into the improved performance of TSNPE as compared to NPE, we computed the fraction of prior samples that lie within the $\text{HPR}_{\epsilon}$ of the approximate posterior (Fig.~\ref{fig:fig4}, fourth column). In tasks with broad posteriors and few simulations, the $\text{HPR}_{\epsilon}$ is almost as wide as the prior and thus the performances of NPE and TSNPE are similar (e.g., SLCP with 1k simulations). In other tasks and with more simulations, the $\text{HPR}_{\epsilon}$ is much narrower than the prior, leading to an improvement in simulation efficiency (e.g., Lotka-Volterra with 100k simulations).

Finally, we evaluated whether the $\text{HPR}_{\epsilon}$ of the approximate posterior contains the support of the true posterior (Fig.~\ref{fig:fig4}, fifth column). We computed the fraction of true-posterior samples within the $\text{HPR}_{\epsilon}$ of the approximate posterior. For most tasks, fewer than 0.1\% of samples were excluded, and the rate of erroneously rejected samples decreased as more simulations were used. In the Lotka-Volterra task with 1k and 10k simulations, many ground-truth samples were rejected and TSNPE performed poorly, but NPE and APT also failed to solve the task. Thus, while truncated proposals can potentially induce posterior biases, these only have a negligible effect on the performance of TSNPE. We note that TSNPE performance is qualitatively unaffected by the choice of $\epsilon \le 10^{-4}$ and proposal sampling scheme (Appendix Fig.~\ref{fig:fig4:supp1}, Fig.~\ref{fig:fig4:supp2}, Fig.\ref{fig:fig4:supp3}). Applying truncated proposals to APT leads to equally good or worse performance than 'standard' APT, depending on the task (Appendix Fig.~\ref{fig:fig4:supp5}).

\subsection{Efficient and robust inference in two complex neuroscience problems}

%We showed that TSNPE achieves state-of-the-art performance on six benchmark tasks.
Next, we evaluate the performance of TSNPE on two challenging neuroscience problems, where the competitive advantage of TSNPE is fully realized.

% \subsection{Failures of APT for neuroscience simulator}
\paragraph{Pyloric network}
\label{sec:results:pyloric}

\begin{figure}[t]
\ifdefined\NOFIGS
    [FIGURE SKIPPED]
    [Comment out line 3 of main.tex to include]
\else
    \includegraphics[width=\linewidth]{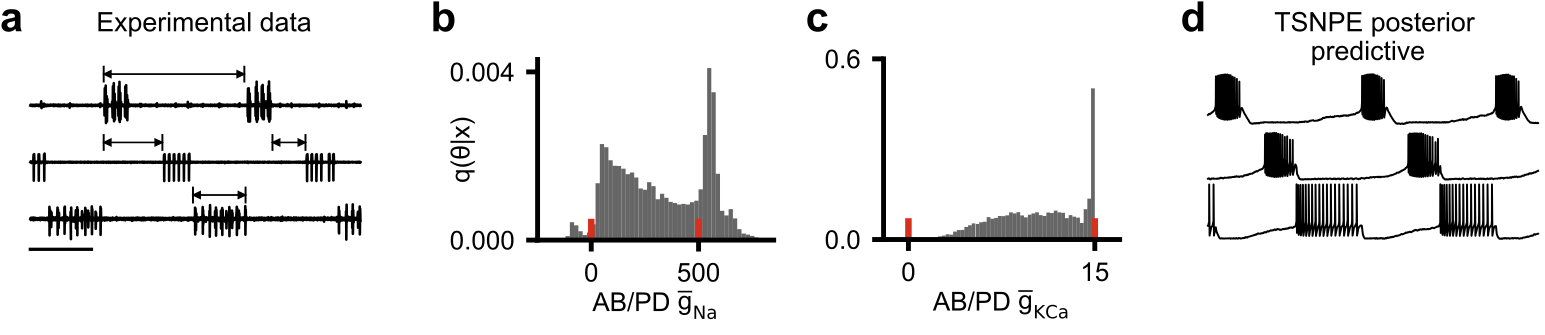}
\fi
\caption{
{\bf Pyloric network inference.}
(a) Data \citep{haddad10recordings}.
(b) APT approximate posterior (1D-marginal) exhibits leakage (red: prior bounds).
(c) APT approximate posterior when forcing the density estimator into constrained space. The spike at the upper prior bound is at odds with previously published posterior distributions \citep{gonccalves2020training, deistler2021disparate, gloeckler2022variational} and produces poor predictive samples (Appendix Fig.~\ref{fig:fig7:supp1}).
(d) TSNPE posterior predictive sample matches summary statistics of the experimental data.
}

\label{fig:fig7}
\end{figure}

We applied TSNPE to a challenging real-world simulator from neuroscience: The pyloric network of the stomatogastric ganglion in the crab \textit{Cancer Borealis} \citep{prinz2003alternative, prinz2004similar}. The model has $31$ parameters and simulates 3 voltage traces that we reduce to $18$ summary statistics. The prior distribution is uniform within previously described parameter ranges \citep{prinz2004similar, gonccalves2020training}. We identify the posterior distribution given experimentally observed data \citep{haddad10recordings} (Fig.~\ref{fig:fig7}a) with APT and TSNPE (13 rounds, 30k simulations per round).

When applying APT `out of the box' (from `sbi' toolbox \citep{tejerocantero2020sbi}), the rate of approximate-posterior samples within the prior bounds was 0.02\% after the second round and 0.0000\% after the third round (Fig.~\ref{fig:fig7}b), which rendered a fourth round too computationally expensive.

We attempted to overcome these issues by appending a transformation $T$ to the density estimator $q_{\bphi}(\btheta | \bx)$ such that its support is constrained to match the support of the prior. In practice, we used a sigmoid transformation. While the resulting approximate posterior exhibited no leakage, this setup revealed another problem when running APT: In transformed (i.e., unbounded) space, the density estimator $q_{\bphi}(\btheta | \bx)$ can put significant mass in regions outside of the training data. When forced into constrained space, these `leaking' regions lead to spikes at the bounds of the parameter space (Fig.~\ref{fig:fig7}c, further details in Appendix Sec.~\ref{sec:appendix:apt}; illustration, additional tests and full posterior in Appendix Figs.~\ref{fig:fig7:supp5}, \ref{fig:fig7:supp4} and \ref{fig:fig7:supp6}). These spikes are at odds with previously published posterior distributions \citep{gonccalves2020training, deistler2021disparate, gloeckler2022variational} and samples from these parameter regions do not produce good simulations (Appendix Fig.~\ref{fig:fig7:supp1}). This demonstrates that leakage occurs in APT even when the density estimator is forced into constrained space and that these issues lead to an incorrect posterior approximation as well as to poor predictive samples.

We applied TSNPE to this task for 13 rounds without any issue. The resulting posterior produces samples that closely match the observed data (Fig.~\ref{fig:fig7}d, more samples in Appendix Fig.~\ref{fig:fig7:supp2}, posterior distribution across all $31$ parameters in Appendix Fig.~\ref{fig:fig7:supp3}). The obtained posterior is similar to previously published posteriors \citep{gonccalves2020training, deistler2021disparate, gloeckler2022variational}.

\begin{figure}[t]
\ifdefined\NOFIGS
    [FIGURE SKIPPED]
    [Comment out line 3 of main.tex to include]
\else
    \includegraphics[width=\linewidth]{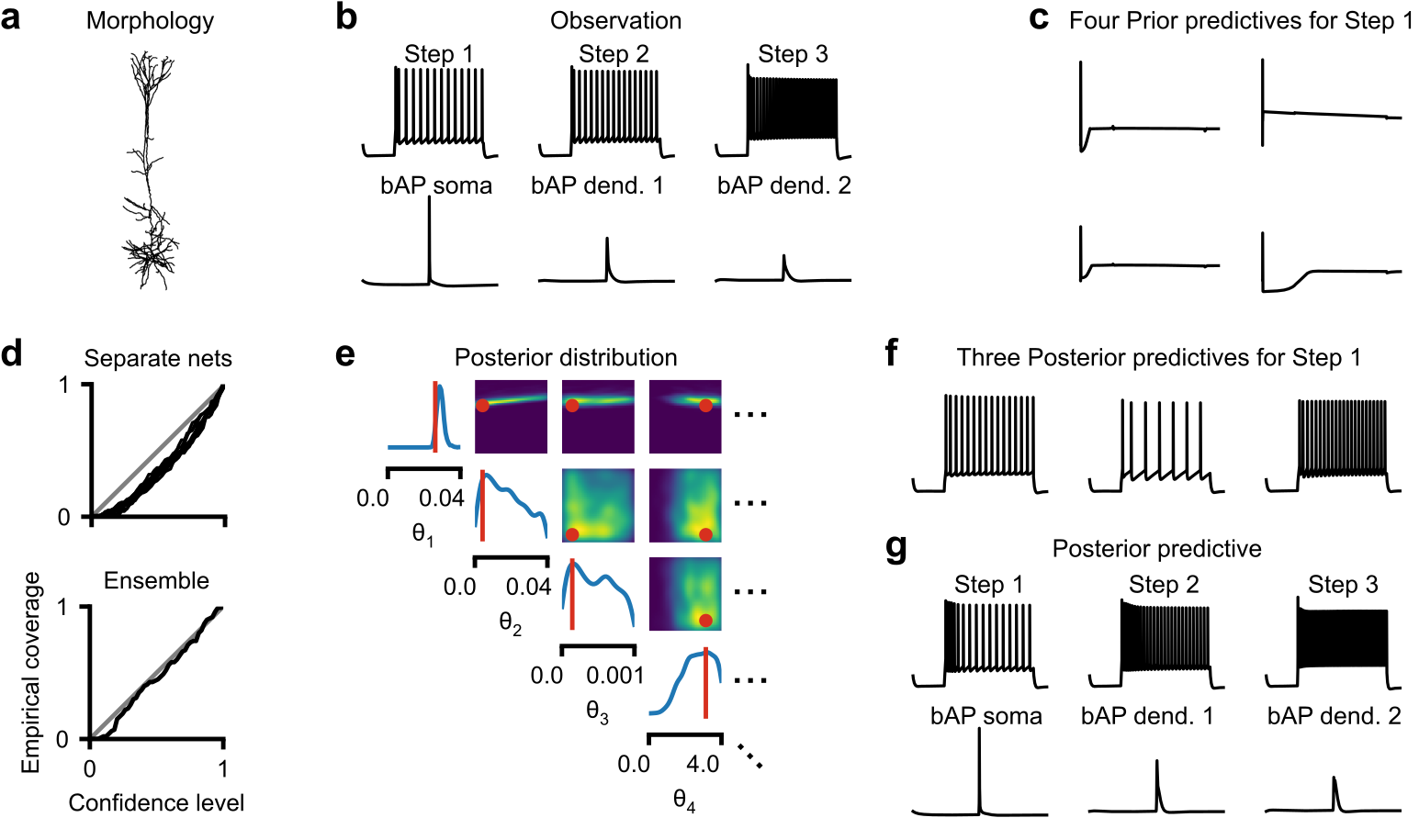}
\fi
\caption{
{\bf TSNPE on L5PC.}
(a) Cell morphology.
(b) Observation.
(c) Four prior samples for the Step 1 protocol.
(d) Coverage for 1 and 10 neural nets.
(e) Posterior. True parameter in red.
(f) Three posterior predictives for the Step 1 protocol.
(g) One posterior predictive for all protocols.
}

\label{fig:fig5}
\end{figure}

\begin{figure}[ht!]
\ifdefined\NOFIGS
    [FIGURE SKIPPED]
    [Comment out line 3 of main.tex to include]
\else
    \includegraphics[width=\linewidth]{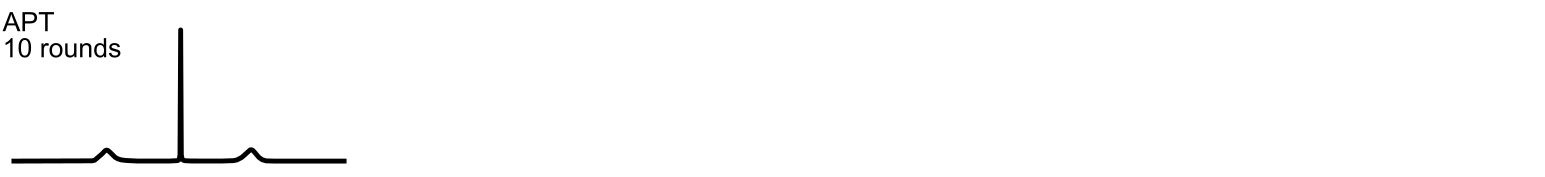}
\fi
\caption{
{\bf APT performance after 10 rounds.}
Same setup as in Fig.~\ref{fig:fig6}, but after running APT for 10 rounds. Leakage gets worse when additional rounds are run.
}

\label{fig:fig6:supp2}
\end{figure}

\paragraph{Multicompartment model of a single neuron}
\label{sec:results:l5pc}
% Model description.
Finally, we turn to a landmark problem in neuroscience for which the posterior has not yet been identified: A morphologically detailed model of a thick-tufted layer 5 pyramidal cell (L5PC) from the neocortex \citep{ramaswamy2015neocortical, markram2015reconstruction, van2016bluepyopt}. The model describes the response of a neuron to current stimuli of different strengths. The model has approximately 7000 separate compartments which compose the anatomy of the cell (Fig.~\ref{fig:fig5}a). Each compartment has dynamics based on the Hodgkin-Huxley equations \citep{hodgkin1952quantitative} and contains multiple ion channels (details in \citet{van2016bluepyopt}). The model has 20 free parameters which are the maximal channel conductances and time constants of the ion channels. The simulation consists of four separate simulations corresponding to experimental protocols which describe the voltage response to different stimuli. For the first three protocols (Step 1, Step 2, Step 3), each voltage response is characterized by 10 summary statistics. The fourth protocol models the back-propagation of the voltage response through the dendritic tree and is captured by 5 additional summary statistics (bAP soma, bAP dend.~1, bAP dend.~2). In total, the model produces 35 summary statistics, to which we add Gaussian noise with diagonal covariance matrix capturing the response variability of previously reported measurements \citep{hay2011models}.

% Difficulty of fitting this model.
Our goal is to infer the posterior distribution over 20 parameters given 35 summary statistics that were simulated---and thus have a known ground-truth parameter set--- to resemble experimentally observed activity (Fig.~\ref{fig:fig5}b). The prior is a uniform distribution within previously established bounds \citep{van2016bluepyopt}. A major difficulty in fitting this model is that a large fraction of prior samples generate summary statistics that are very different from the observed data: In particular, about 99.98\% of prior predictives contain at least one summary statistic that is undefined, e.g., time to first spike is undefined in the absence of spikes (Fig.~\ref{fig:fig5}c). When a summary statistic is undefined, we assign it a value that is substantially outside the range of the observed data (Appendix Sec.~\ref{sec:appendix:l5pcmodel}).

% Our general setup -- describe coverage analysis and tempering.
We ran TSNPE over six rounds (hyperparameters in Appendix Sec.~\ref{sec:appendix:hyperparameters}). In each round, we ran 30k simulations, leading to a total of 180k simulations. After every round, we evaluated the expected coverage with SBCC. After the first round, the approximate posterior exhibited poor expected coverage (Fig.~\ref{fig:fig5}d top). Therefore, as suggested by \citet{hermans2021averting}, we used an ensemble of 10 neural density estimators to ensure that the approximate posterior is sufficiently broad (Fig.~\ref{fig:fig5}d bottom). Although the approximate posterior remains underconfident, the empirical expected coverage closely matches the confidence level for high-confidence levels, which is crucial for TSNPE (Sec.~\ref{sec:methodology:coverage}).

The TSNPE-posterior has several parameters with broad marginals, demonstrating that this model exhibits `degeneracy', a widespread phenomenon in biological systems \citep{marder2011multiple} (Fig.~\ref{fig:fig5}e, Appendix Sec.~\ref{sec:appendix:l5pcmodel} for parameter names). Other marginals are narrow, demonstrating that the model is sensitive to changes in these parameters. Posterior predictive samples closely match the observed data (Fig.~\ref{fig:fig5}f,g; more samples in Appendix Fig.~\ref{fig:fig5:supp2}). We emphasize that fitting such morphologically-detailed neuron models is a challenging and widespread problem in neuroscience, one for which commonly used methods \cite[e.g., genetic algorithms, ][]{Druckmann_07,van2016bluepyopt} are often simulation-inefficient or do not estimate the full posterior distribution. We show promising results, suggesting that TSNPE could be applied to other complex single-neuron models.

In contrast, with `out of the box' APT, none of the 10M approximate-posterior samples was within the prior bounds after the second round. This rendered a third round too computationally expensive. Overall, the results on the pyloric network model and on the multicompartment model demonstrate that TSNPE is an efficient and robust method that scales to complex and high-dimensional models that were inaccessible to the state-of-the-art method APT.

\section{Discussion}
\label{sec:discussion}

% Summary
We presented a new method to perform Bayesian inference in implicit models, which we call Truncated Sequential Neural Posterior Estimation (TSNPE). Like previous methods, TSNPE adaptively selects parameters to improve simulation-efficiency and allow posterior inference in complex models with many parameters. The key ingredient is that TSNPE samples parameters from a `truncated' region of the prior, and thus overcomes instabilities of previous methods while maintaining simulation efficiency. In order to diagnose potential errors of TSNPE, we developed a coverage test that can be run quickly and at every round of TSNPE. TSNPE presents a new variant of SNPE which is at least as powerful as previous variants on benchmark tasks, but provides a powerful alternative which is able to solve inference problems on which the state-of-the-art method APT failed.

\paragraph{Related work}
\label{sec:discussion:related}

TSNPE differs from automatic posterior transformation (APT, SNPE-C) in its proposal and its loss function: TSNPE uses a truncated prior as proposal, while APT can flexibly use any proposal, which, e.g.,  allows for more sophisticated active learning rules \citep{lueckmann2019likelihood, jarvenpaa2019efficient}. However, APT's flexibility requires a modification of NPE loss function, which can be an impediment to its usage in practice: First, the modification can lead to `leakage', which can make it prohibitive to draw samples within prior bounds. Second, APT loss requires an explicit prior and thus, cannot be applied to models in which the prior can only be sampled \citep{ramesh2022gatsbi}. Third, current formulations of APT cannot discard parameters leading to invalid simulations as the posterior mass would `leak' into parameter regions which only produce invalid simulations (Appendix Sec.~\ref{sec:appendix:apt}). It might be possible that these issues are resolved using a modified formulation of APT, e.g., by combining its atomic loss with additional loss terms, or preventing leakage by penalizing `bad' parameters \citep{greenberg2019automatic}. In cases in which leakage prevents application of APT (in particular, in high-dimensional problems), TSNPE provides an alternative.

% Comparison to Blum Francois
Our method is inspired by previous work that introduced a mechanism to post-hoc correct samples obtained by an Approximate Bayesian Computation (ABC) algorithm \citep{BlumFrancois_10}, i.e., `regression adjustment ABC'. Their method draws samples from a truncated region of the prior to avoid correction terms, but estimates the posterior density with ABC samples---rather than using a flexible neural density estimator---and estimates the support by training a dedicated support-vector machine. In addition, the method runs a single round of truncation and retraining, whereas we demonstrate that TSNPE can be robustly applied across 10 rounds.

% Comparison to TMNRE.
Truncated proposals have also been proposed for neural ratio estimation \citep[Truncated Marginal Neural Ratio Estimation (TMNRE)][]{miller2021truncated}: TMNRE uses truncated proposals to efficiently infer selected posterior marginals while being amortised around the observation, allowing to test the coverage properties of the selected marginals, e.g., with SBC \citep{cook2006validation,talts2018validating}. In addition, truncating based on the marginals allows TMRNE to sample from the truncated proposal without rejection or SIR sampling. In contrast, TSNPE aims at efficiently inferring the full posterior distribution by proposals that avoid the correction of SNPE loss function. Truncating the proposal based on the full posterior rather than on the marginals can lead to drastically narrower proposals: E.g., on the pyloric network problem, truncation based on posterior marginals rejects 20\% of prior samples versus 99.94\% rejection based on the posterior joint.
% While the authors of TMNRE state that their truncation algorithm could be extended to truncating the full posterior with nested sampling, they did not explore this avenue.
In addition, while TMNRE uses the expected coverage to test the consistency of the posterior marginals, TSNPE can test the expected coverage of the full posterior distribution.

\paragraph{Possible failure modes}
\label{sec:discussion:failures}

The main failure mode of TSNPE will occur if the truncated proposal excludes significant portions of density mass of the true posterior (e.g., if the estimate misses posterior modes). In these cases, the learned approximate posterior will put systematically too little mass in the excluded regions. We recommend the use of diagnostic tools such as SBCC to identify such failures \citep{cook2006validation, miller2021truncated, hermans2021averting, rozet2021arbitrary}.

In addition, if the true posterior has unbounded support, any finite values of $\epsilon > 0$ will lead to a biased approximate posterior which puts too little weight in the posterior tails. In that case, and when running TSNPE across many rounds, the errors from each individual round could accumulate. Although we did not observe this bias to significantly affect the algorithm performance on several benchmark tasks, we cannot exclude the possibility of a substantial performance degradation when running TSNPE for a larger number of rounds ($\gg$10).

Finally, unlike SNPE methods that use the previous estimate of the posterior as the proposal distribution, our method requires a scheme to sample from a truncated proposal. If the sampling scheme is inaccurate (i.e., if it does not produce a proposal distribution that is proportional to the prior within the truncated region), the results of TSNPE will be biased. To avoid this, we recommend using rejection sampling by default and using SIR or sequential Monte-Carlo methods only if rejection sampling is too computationally expensive. For SIR, it is important to use a large oversampling factor $K$ (e.g., $K=1024$) and use diagnostic tools such as effective sample size (Appendix Sec.~\ref{sec:appendix:sampling_sir_accuracy}, Fig.~\ref{fig:fig4:supp6}).

% New test for diagnosis.
\paragraph{Simulation-based coverage calibration}
\label{sec:discussion:diagnostics}

In order to diagnose whether the approximate posterior is broader than the true posterior, we applied SBCC, a coverage test for TSNPE \citep{cook2006validation, rozet2021arbitrary}. SBCC evaluates the expected coverage of the approximate posterior without evaluating it on a grid \citep{dalmasso2020conditional, hermans2021averting} and, unlike diagnostic tools for methods based on learning the likelihood(-ratio), does not require MCMC runs for multiple observations $\bx$ \citep{miller2021truncated}. This allows SBCC to be run quickly and for models with many parameters. 
In addition, in contrast to diagnostic tools for likelihood-free inference with Approximate Bayesian Computation, SBCC does not require an additional step to estimate the density of approximate posterior samples \citep{prangle2014diagnostic}. We note that, since SBCC is a variation of SBC \citep{cook2006validation, talts2018validating}, it only ensures that the $\text{HPR}_{\epsilon}$ is correct \emph{on average} across observations, not for a particular observation. In principle, SBCC could be applied to other SNPE variants, although empirically the impact of arbitrary proposals on SBCC performance is currently unclear.

\paragraph{Conclusion}
\label{sec:discussion:conclusion}
Overall, TSNPE combines the simulation-efficiency of sequential neural posterior estimation with the robustness and coverage-tests of non-sequential methods. We demonstrated that it 
%can be applied to a wide range of problems and thereby 
allows to scale neural posterior estimation to complex and high-dimensional scientific problems.

\begin{ack}
We thank Poornima Ramesh, Cornelius Schr{\"o}der, Marcel Nonnenmacher, David Greenberg, and Jan-Matthis Lueckmann for discussions and feedback. We also thank the International Max Planck Research School for Intelligent Systems (IMPRS-IS) for supporting MD. This work was funded by the German Research Foundation (DFG; Germany’s Excellence Strategy MLCoE – EXC number 2064/1 PN 390727645) and the German Federal Ministry of Education and Research (BMBF; Tübingen AI Center, FKZ: 01IS18039A).
\end{ack}

\bibliography{references}

\newpage
%%%%%%%%%%%%%%%%%%%%%%%%%%%%%%%%%%%%%%%%%%%%%%%%%%%%%%%%%%%%

\section*{Checklist}

%%% BEGIN INSTRUCTIONS %%%
% The checklist follows the references.  Please
% read the checklist guidelines carefully for information on how to answer these
% questions.  For each question, change the default \answerTODO{} to \answerYes{},
% \answerNo{}, or \answerNA{}.  You are strongly encouraged to include a {\bf
% justification to your answer}, either by referencing the appropriate section of
% your paper or providing a brief inline description.  For example:
% \begin{itemize}
%   \item Did you include the license to the code and datasets? \answerYes{See Section~\ref{gen_inst}.}
%   \item Did you include the license to the code and datasets? \answerNo{The code and the data are proprietary.}
%   \item Did you include the license to the code and datasets? \answerNA{}
% \end{itemize}
% Please do not modify the questions and only use the provided macros for your
% answers.  Note that the Checklist section does not count towards the page
% limit.  In your paper, please delete this instructions block and only keep the
% Checklist section heading above along with the questions/answers below.
%%% END INSTRUCTIONS %%%

\begin{enumerate}

\item For all authors...
\begin{enumerate}
  \item Do the main claims made in the abstract and introduction accurately reflect the paper's contributions and scope? \answerYes{}
  \item Did you describe the limitations of your work? \answerYes{}
  \item Did you discuss any potential negative societal impacts of your work? \answerNA{}
  \item Have you read the ethics review guidelines and ensured that your paper conforms to them? \answerYes{}
\end{enumerate}

\item If you are including theoretical results...
\begin{enumerate}
  \item Did you state the full set of assumptions of all theoretical results? \answerYes{}
        \item Did you include complete proofs of all theoretical results? \answerYes{}
\end{enumerate}

\item If you ran experiments...
\begin{enumerate}
  \item Did you include the code, data, and instructions needed to reproduce the main experimental results (either in the supplemental material or as a URL)?
    \answerYes{}
  \item Did you specify all the training details (e.g., data splits, hyperparameters, how they were chosen)? \answerYes
  \item Did you report error bars (e.g., with respect to the random seed after running experiments multiple times)? \answerYes
  \item Did you include the total amount of compute and the type of resources used (e.g., type of GPUs, internal cluster, or cloud provider)? \answerYes{}
\end{enumerate}

\item If you are using existing assets (e.g., code, data, models) or curating/releasing new assets...
\begin{enumerate}
  \item If your work uses existing assets, did you cite the creators? \answerYes{}
  \item Did you mention the license of the assets? \answerNA{}
  \item Did you include any new assets either in the supplemental material or as a URL? \answerNA{}
  \item Did you discuss whether and how consent was obtained from people whose data you're using/curating? \answerNA{}
  \item Did you discuss whether the data you are using/curating contains personally identifiable information or offensive content? \answerNA{}
\end{enumerate}

\item If you used crowdsourcing or conducted research with human subjects...
\begin{enumerate}
  \item Did you include the full text of instructions given to participants and screenshots, if applicable?
    \answerNA{}
  \item Did you describe any potential participant risks, with links to Institutional Review Board (IRB) approvals, if applicable?
    \answerNA{}
  \item Did you include the estimated hourly wage paid to participants and the total amount spent on participant compensation?
    \answerNA{}
\end{enumerate}

\end{enumerate}

%%%%%%%%%%%%%%%%%%%%%%%%%%%%%%%%%%%%%%%%%%%%%%%%%%%%%%%%%%%%

\newpage
\section{Appendix}
\label{sec:appendix}

\subsection{Reproducibility statement}
We used the configuration manager hydra to track the configuration and seeds of each run \citep{Yadan2019Hydra}. All code to reproduce the results can be found at \url{https://github.com/mackelab/tsnpe_neurips}. We implemented TSNPE on top of the publicly accessible sbi toolbox \citep{tejerocantero2020sbi}. All simulations and runs were performed on a high-performance computer. For each run, we used between 8 and 48 CPU cores.

\subsection{Proof of convergence}
\label{sec:appendix:proofconvergence}

Below, we prove that, for a given observation $\bx_o$, the posterior distribution obtained with TSNPE $q_{\phi}(\btheta | \bx_o)$ converges to the true posterior distribution $p(\btheta | \bx_o)$ under the assumption that the $\text{HPR}_{\epsilon}$ of the approximate posterior at every TSNPE round covers the support (i.e., the $\text{HPR}_{\epsilon}$ for $\epsilon=0$) of the true posterior given the observation $\bx_o$. We call the support of the true posterior $\mathcal{M}$, i.e.~$\mathcal{M} = \text{supp}(p(\btheta | \bx_o))$. The proof proceeds in two steps: First, we derive the effective proposal distribution when pooling data from all rounds. Second, we show that, for such proposal distributions, the neural density estimator converges to the true posterior.

\paragraph{Deriving the proposal distribution}
We denote by $\tilde{p}^r(\btheta)$ the proposal distribution from which $\btheta$ are drawn in round $r$. In the first round, we use the prior, i.e., $\tilde{p}^{r=1}(\btheta) = p(\btheta)$. In later rounds, we sample from the prior but reject samples that lie outside of the $\text{HPR}_{\epsilon}$ of the approximate posterior given $\bx_o$. The samples drawn in round $r$ are thus drawn from
\begin{equation}
    \tilde{p}^r(\btheta) = U^r(\btheta) p(\btheta) / Z^r,
\end{equation}
where $Z^r$ is the normalization constant and $U^r(\btheta)$ is $1$ on the $\text{HPR}_{\epsilon}$ of the approximate posterior at round $r$ and zero otherwise. Given our assumption that the $\text{HPR}_{\epsilon}$ of the approximate posterior covers the support of the true posterior $\mathcal{M}$, $U^r(\btheta)$ is $1$ on the support of the true posterior $\mathcal{M}$.

When pooling simulations from all rounds, after $R$ rounds, the parameters are sampled from a mixture of all proposal distributions:
\begin{equation}
    \tilde{p}(\btheta) = \frac{1}{N} \sum_{r=1}^R \tilde{p}^r(\btheta) = p(\btheta) \cdot \Bigg( \frac{1}{N} \sum_{r=1}^R U^r(\btheta) / Z^r \Bigg) = p(\btheta) \cdot f(\btheta).
\end{equation}
In this equation, we assumed that all rounds contain equally many simulations, but the proof can easily be extended to rounds with different numbers of simulations by adding weights to the above sum.

As can be seen above, the distribution $\tilde{p}(\btheta)$ is the prior times a function $f(\btheta)$ which is made up of several steps and whose steps are defined by the $\text{HPR}_{\epsilon}$ of the approximate posterior of every round (illustration in Appendix Fig.~\ref{fig:fig6:supp1}). Finally, under the assumption that all $U^r(\btheta)$ are $1$ on the support of the true posterior $\mathcal{M}$, we have, for any $\btheta \in \mathcal{M}$
\begin{equation*}
    f(\btheta) = \frac{1}{N} \sum_{r=1}^R U^r(\btheta) / Z^r_1 = \frac{1}{N} \sum_{r=1}^R 1/Z^r_1 = \text{constant} = c,
\end{equation*}
i.e., $f(\btheta)$ is not a function of $\btheta$. Thus, for any $\btheta \in \mathcal{M}$, we have
\begin{equation*}
    \tilde{p}(\btheta) = p(\btheta) \cdot c \propto p(\btheta)
\end{equation*}
We emphasise that this proportionality holds only for $\btheta \in \mathcal{M}$, but not necessarily for $\btheta$ outside of the support of the true posterior.

\paragraph{Training the neural density estimator}
Next, we show that, for a proposal distribution of the form derived in the paragraph above, the approximate posterior $q_{\phi}(\btheta | \bx_o)$ for an observation $\bx_o$ converges to the true posterior distribution $p(\btheta | \bx_o)$.

TSNPE minimizes the following loss function:
\begin{equation*}
\begin{split}
    \mathcal{L} & = -\frac{1}{N} \sum_i \log(q_{\bphi}(\btheta_i|\bx_i)) \\
    & \xrightarrow{N \rightarrow \infty} -\EX_{p(\btheta,\bx)}[\log(q_{\bphi}(\btheta|\bx))] \\
    & = -\EX_{\tilde{p}(\btheta)p(\bx|\btheta)}[\log(q_{\bphi}(\btheta|\bx))].
\end{split}
\end{equation*}
Plugging in the proposal distribution defined above:
\begin{equation*}
\begin{split}
    \mathcal{L} & = -\iint f(\btheta)p(\btheta)p(\bx|\btheta)  \log(q_{\phi}(\btheta|\bx)) \; d\btheta d\bx \\
    & = \int p(\bx) \int -f(\btheta)p(\btheta|\bx) \log(q_{\phi}(\btheta|\bx)) \; d\btheta d\bx.
\end{split}
\end{equation*}
The term within the integral over $\btheta$ is proportional to the Kullback-Leibler-divergence between $f(\btheta)p(\btheta|\bx)/Z$ (with $Z=\int f(\btheta)p(\btheta|\bx) d\btheta$) and the approximate posterior $q_{\phi}(\btheta|\bx)$. Thus, $\mathcal{L}$ is minimized if and only if 
\begin{equation*}
    q_{\phi}(\btheta|\bx) \propto f(\btheta)p(\btheta|\bx)
\end{equation*}
for all $\bx$ within the the support of $p(\bx)$ \citep{papamakarios2016fast}.

This means that, for arbitrary $\bx \in \text{supp}(p(\bx))$, $q_{\phi}(\btheta | \bx)$ will not converge to the true posterior $p(\btheta | \bx)$, but to $f(\btheta)p(\btheta|\bx)/Z$. However, for the observed data $\bx = \bx_o$, we have:
\begin{equation*}
    q_{\phi}(\btheta|\bx_o) \propto f(\btheta)p(\btheta|\bx_o) =
    \begin{cases}
        c \cdot p(\btheta|\bx_o) & \text{if } \btheta \in \mathcal{M} \\
        f(\btheta) \cdot 0 & \text{else}
    \end{cases}
\end{equation*}
The first case follows from the fact that $f(\btheta)$ is constant on the support of the true posterior. The second case follows because the true posterior has zero probability density for $\btheta$ outside of its own support $\mathcal{M}$. Thus:
\begin{equation*}
    q_{\phi}(\btheta|\bx_o) \propto p(\btheta|\bx_o).
\end{equation*}
Since $q_{\phi}(\btheta|\bx_o)$ is a (conditional) normalizing flow, it is normalized and, thus:
\begin{equation*}
    q_{\phi}(\btheta|\bx_o) = p(\btheta|\bx_o).
\end{equation*}

\subsection{Simulation-based coverage calibration (SBCC)}
\label{sec:appendix:algorithmcoverage}

The algorithm for computing the coverage (SBCC) is shown in Alg.~\ref{alg:alg2}.

\begin{algorithm}[ht]
 \textbf{Inputs:} proposal $\Tilde{p}(\btheta)$, approximate posterior $q_{\bphi}$ trained on $\btheta \sim \Tilde{p}(\btheta)$, number of simulations $M$, number of posterior samples per simulation $P$. Database of empirical coverages $\mathcal{E}$, initialized as the empty set.
 
 \textbf{Outputs:} Coverage for different confidence levels $1-\alpha$.
 
\For{$i \in [1, ..., M]$}{
    $\btheta_i^* \sim \Tilde{p}(\btheta)$ \\
    simulate $\bx_i^* \sim p(\bx|\btheta_i^*)$\\
    $l^*_i = \log(q_{\bphi}(\btheta_i^* | \bx_i^*))$  \tcp*{compute log-prob of ground-truth}
    $c = 0$\\
    \For{$j \in [1, ..., P]$}{
        $\btheta_j \sim q_{\bphi}(\btheta | \bx_i^*)$\\
        $l_j = \log(q_{\bphi}(\btheta_j | \bx_i^*))$  \tcp*{compute log-probs of posterior samples}
        \If{$l_j > l^*_i$}{
            $c = c + 1$
        }
    
        $e = c / P$ \tcp*{fraction of posterior samples whose log-prob is larger \newline than ground-truth log-prob} 
        
        add $e$ to $\mathcal{E}$
    }
}

Plot CDF of $\mathcal{E}$

\caption{Simulation-based coverage calibration (SBCC)}
\label{alg:alg2}
\end{algorithm}

\subsection{Algorithm description for sampling from the HPR of the approximate posterior}
\label{sec:appendix:algorithmsampling}
The algorithm for sampling from the $\text{HPR}_{\epsilon}$ of the approximate posterior is shown in Alg.~\ref{alg:alg3}.

\begin{algorithm}[ht]
 \textbf{Inputs:} Prior $p(\btheta)$, Approximate posterior $q_{\bphi}$, observation $\bx_o$, number of posterior samples $M$, $\epsilon$ that defines the HPR, number of desired samples from $\tilde{p}(\theta)$ $N$. Database of log-probabilities $\mathcal{P}$, initialized as an empty set.
 
 \textbf{Outputs:} Samples $\mathcal{S}$ from truncated proposal $\tilde{p}(\btheta)$.
 
\For{$i \in [1, ..., M]$}{
    $\btheta_i \sim q_{\bphi}(\btheta | \bx_o)$ \\
    $l_i = \log(q_{\bphi}(\btheta_i | \bx_o)$\\
    add $l_i$ to $\mathcal{P}$
}

$\bkappa$ = quantile$_{\epsilon}(\mathcal{P})$ \tcp*{threshold for $\text{HPR}_{\epsilon}$}
~\\
\tcp{sample truncated proposal with rejection sampling}
\While{$s < N$}{ 
    $\btheta \sim p(\btheta)$ \\
    $l = \log(q_{\bphi}(\btheta | \bx_o))$\\
    \If{$l > \bkappa$}{
        add $\btheta$ to $\mathcal{S}$\\
        s += 1
    }
}

\caption{Obtaining the $\text{HPR}_{\epsilon}$ of the approximate posterior and sampling the truncated proposal with rejection sampling}
\label{alg:alg3}
\end{algorithm}

\subsection{Alleviating issues of APT}
\label{sec:appendix:apt}
We compared TSNPE to Automatic Posterior Transformation (APT) \citep{greenberg2019automatic}. In this section, we briefly review how APT works, and why `leakage' occurs, and how we attempted to improve it. % (despite not fixing all of its issues, see Sec.~\ref{sec:results:pyloric}).

\paragraph{APT review} Broadly, APT exists in two versions. Its first version can be applied only if the density estimator $q_{\bphi}(\btheta | \bx)$ is a mixture of Gaussians, the proposal $\tilde{p}(\btheta)$ is a mixture of Gaussians, and the prior $p(\btheta)$ is either uniform or Gaussian. We did not compare TSNPE to APT in this form because we wanted to use more expressive density estimators for $q_{\bphi}(\btheta | \bx)$. The second version of APT, known as atomic APT, allows to use any density estimator $q_{\bphi}(\btheta | \bx)$, any proposal $\tilde{p}(\btheta)$ and any explicit prior $p(\btheta)$. One must be able to evaluate the density estimator and the prior, but the proposal can be implicit (i.e., without closed form density). Atomic APT minimizes the loss:
\begin{equation*}
    \mathcal{L}_{\bphi} = - \frac{1}{N} \sum_{i=1}^N \log \frac{q_{\bphi}(\btheta_i | \bx_i) / p(\btheta_i)}{q_{\bphi}(\btheta_i | \bx_i) / p(\btheta_i) + \sum_{j=1...A-1} q_{\bphi}(\btheta_j | \bx_i) / p(\btheta_j)}
\end{equation*}
with number of atoms $A$. In this loss, $\btheta_j$ and $\btheta_i$ can be sampled from \emph{any} proposal distribution. $\bx_i$ is sampled from $p(\bx | \btheta_i)$.

\paragraph{Leakage issue}
Notice that the above loss is the same for $q_{\bphi}(\btheta | \bx)$ and for $c \cdot q_{\bphi}(\btheta | \bx)$. In other words, the approximate posterior has to be correct only up to a proportionality constant \citep{greenberg2019automatic, durkan2020contrastive}. Thus, an approximate posterior $q_{\bphi}(\btheta | \bx)$ which is $c \cdot p(\btheta | \bx)$ will have a minimal loss. Since the dataset on which $q_{\bphi}(\btheta | \bx)$ is trained contains no $\btheta$ that lies outside of the prior bounds, the approximate posterior can be anything outside of the prior bounds without affecting the value of the loss function. This is what is called `leakage', and which has been pointed out as a potential problem both in the original APT paper \cite{greenberg2019automatic} and work studying the relationship of APT with contrastive learning approaches \cite{durkan2020contrastive}: While the approximate posterior might be proportional to the true posterior within the bounds of the prior $p(\btheta)$, it can put significant mass outside of the prior bounds (because the loss does not penalize this behaviour).

\paragraph{Transforming the parameter space}
We tried to fix the leakage issue by appending a transformation such that the density estimator has constrained support. This requires that the bounds of the parameter space are known and that such a transformation can be implemented. If this fix can be applied, the leakage will be zero by definition. In all our experiments, such a transformation substantially helped with the leakage problem.

\paragraph{Leakage into regions of no training data}
However, even when possible, such transformation does not completely prevent `leakage': When inspecting the loss of APT, we see that the approximate posterior can put mass into any region of the parameter space in which no parameter sets $\btheta$ in the training data lie. For example, if the prior distribution is standard Gaussian and one trains APT on $100$ parameter sets sampled from the prior, the training dataset will unlikely contain values that are smaller than -3 or larger than 3. Therefore, APT can `leak' into the regions below -3 or above 3 and still have an optimal loss. While, in the limit of infinite data, APT would correct its `leakage' as soon as parameters from these regions are used as training data, given a finite number of simulations, the approximate posterior can `leak' into new regions which have not been explored (yet). The (potential) problem of leakage does not affect all applications equally: High-dimensional parameter spaces suffer more from this behavior than low-dimensional ones, as there are many regions into which the mass of the approximate posterior can `leak'. This behaviour is illustrated in Appendix Fig.~\ref{fig:fig7:supp5}.

\paragraph{Leakage because of invalid data}
Another way in which leakage can occur is if the simulator produces invalid data (e.g.~NaN or infinity). Often, such invalid simulations are discarded from the training dataset \citep{lueckmann2017flexible} and the approximate posterior is trained only on samples that produce valid outputs. For example, assume that a small region of the prior always produces invalid simulations. In this case, the approximate posterior will never be trained on simulations from this parameter region and APT can `leak' into this region. Thus, the approximate posterior might contain significant mass in parameter regions that produce invalid simulations.

\paragraph{Explicit recommendations for running APT}
We will now give explicit recommendations for running APT. These modification greatly improved the performance of APT in our experiments, but were not able to avoid the failure of the algorithm on challenging real-world problems such as the pyloric network (Fig.~\ref{fig:fig7}).
\begin{enumerate}
    \item For priors with bounded supports: If possible, transform the parameter space into unbounded space.
    \item Do not discard invalid simulations. Instead, replace invalid entries (such as NaN) with a substantially different value than the observed data and train on all available simulations.
    \item Check if the posterior contains a lot of mass in regions with very low prior probability. If this is the case, it can hint at a failure of APT through leakage into regions of no training data.
    \item If you transformed the parameter space, check if many posterior samples lie very close to the bounds. Again, this can hint at a failure of APT through leakage into regions of no training data.
\end{enumerate}

\subsection{Relation between simulation-based calibration and our diagnostic}
\label{sec:appendix:sbcrelation}

As discussed above, our method is closely related to simulation-based calibration (SBC) \citep{cook2006validation, talts2018validating}. Briefly, SBC samples $\btheta^*$ from the prior, samples the likelihood $\bx^* \sim p(\bx|\btheta^*)$ and then draws samples from the posterior $\btheta_i \sim p(\btheta | \bx^*)$ (e.g., with MCMC). It then projects the (potentially high-dimensional) parameters $\btheta_i$ into a one-dimensional space $T(\btheta_i)$. Often, this projection is the 1D-marginal distribution of parameters \citep{carpenter2017stan}. It then ranks $T(\btheta^*)$ under all posterior samples $T(\btheta_{1...N})$. Repeated across several prior samples, the distribution of ranks should be uniform. For high-dimensional parameter spaces, the marginal distribution of each parameter is checked independently.

Notably, other projections into a one-dimensional space are possible. Below, we explain that our method is identical to running SBC with projection $T: \btheta \rightarrow q_{\bphi}(\btheta | \bx)$ and with posterior samples exactly following $q_{\bphi}(\btheta | \bx)$.% Importantly, these assumptions hold for (T)SNPE, but not for other methods that rely on MCMC sampling.

In SBC and in our diagnostic method, samples are drawn from the prior $\btheta^* \sim p(\btheta)$ and simulated $\bx^* \sim p(\bx | \btheta)$. In our diagnostic method, as well as in SBC with projection being the log-probability of the approximate posterior, one then samples the posterior to obtain $\btheta_i$ and evaluates the log-probability of all samples, i.e., $l_i = T(\btheta_i) = \log(q(\btheta_i | \bx^*)$ as well as of the initial parameter set $l^* = \log(q(\btheta^* | \bx^*)$. SBC then ranks $l^*$ under all $l_i$, which is equivalent to computing its quantile (as in our method). In our diagnostic tool, one then evaluates whether this quantile is above or below several confidence levels (evaluation on a 1D evenly spaced grid, same as rank binning in SBC). This generates a step-function with the step occuring at the quantile of $l^*$. This step function is the cumulative distribution function of a dirac at the quantile of $l^*$. Therefore, repeated across several $\btheta^*$, our coverage plots (e.g., Fig~\ref{fig:fig2}c) correspond to the cumulative distribution function of the histograms generated by SBC (with projection $T: \btheta \rightarrow \log(q_{\bphi}(\btheta | \bx))$ and posterior samples exactly following $q_{\bphi}(\btheta | \bx)$).

\subsection{SBCC in a multi-round setting}
\label{sec:appendix:coverageMultiRound}
We run our diagnostic tool after every round of training. If one trains only on simulations that were run in the most recent round, SBCC can be run as in the first round. However, if one wishes to train on simulations from all rounds, then the deep neural density estimator converges to:
\begin{equation*}
    q_{\phi}(\btheta | \bx) \propto f(\btheta) p(\btheta | \bx).
\end{equation*}
Proof in Sec.~\ref{sec:appendix:proofconvergence}. As is described in Sec.~\ref{sec:appendix:proofconvergence}, this means that $q_{\phi}(\btheta | \bx)$ will not converge to the true posterior for arbitrary $\bx$, but only for the observation $\bx_o$.

This poses a problem for SBCC: As described in Alg.~\ref{alg:alg2}, SBCC measures whether the coverage is correct (on average) for many $\bx$ generated by the proposal distribution. Since the loss employed by TSNPE only ensures convergence for $\bx_o$, it will by construction not provide correct results for other $\bx$.

This issue can be solved in two ways:
\begin{enumerate}
    \item When running SBCC, instead of drawing samples from the proposal prior of the most recent round $\tilde{p}^r(\btheta)$, one can draw samples $\btheta^*$ from $\tilde{p}(\btheta)$, i.e., the distribution that emerges from pooling data from all rounds (notation as in Sec.~\ref{sec:appendix:proofconvergence} and Alg.~\ref{alg:alg2}). This is the method described in Appendix Alg.~\ref{alg:alg2}.
    
    \item One can truncate the approximate posteriors $q_{\phi}(\btheta | \bx^*)$ while running SBCC (see Alg.~\ref{alg:alg2} for notation). With this strategy, when running SBCC in round $r$, we draw parameters from $\btheta^* \sim \tilde{p}^r(\btheta)$ (the truncated proposal from round $r$), simulate them $\bx^* \sim p(\bx | \btheta^*)$, sample from the posterior $\btheta_i \sim q_{\bphi}(\btheta | \bx^*)$, reject samples that lie outside of $\text{HPR}_{\epsilon}(q_{\bphi}(\btheta_i | \bx_o))$, and then continue as described in Alg.~\ref{alg:alg2}. Strategy 2 only ensures that the posterior regions which lie within $\text{HPR}_{\epsilon}(q_{\bphi}(\btheta | \bx_o))$ are well-calibrated. It does not ensure that the full posterior ($q_{\bphi}(\btheta | \bx^*)$ with $\bx^* \sim p(\bx^*|\btheta^*)$ and $\btheta^* \sim \tilde{p}^r(\btheta)$) is well-calibrated.
\end{enumerate}

\subsection{Toy model}
The toy model shown in Fig.~\ref{fig:fig6} is given by a uniform prior within $[-2, -1]$ and $[1, 2]$. The simulator is $x \sim \theta^2 + \epsilon$, where $\epsilon$ is a Gaussian distribution with mean zero and standard deviation $0.2$. We ran APT \citep{greenberg2019automatic} and TSNPE for 5 rounds with 500 simulations per round. For APT, all hyperparameters are the default values from the sbi package \citep{tejerocantero2020sbi}, but we used a neural spline flow (NSF) for both APT and TSNPE \citep{durkan2019neural}.

\subsection{Benchmark tasks}
\label{sec:appendix:benchmark}
Below, we briefly describe the benchmark tasks. For details, please see \citet{lueckmann2021benchmarking}.

\textbf{Gaussian linear:} 10 parameters which are the mean of a Gaussian model. The prior is Gaussian, resulting in a Gaussian posterior.

\textbf{Bernoulli GLM:} Generalized linear model with Bernoulli observations. Inference is performed on 10-dimensional sufficient summary statistics of the originally 100 dimensional raw data. The resulting posterior is 10-dimensional, unimodal, and concave.

\textbf{Lotka Volterra:} A traditional model in ecology \citep{wangersky1978lotka}, which describes a predator-prey interaction between species, illustrating a task with complex likelihood and unimodal posterior.

\textbf{SLCP:} A task introduced by \citet{papamakarios2019sequential} with a simple likelihood and complex posterior. The prior is uniform, the likelihood has Gaussian noise but is non-linearly related to the parameters, resulting in a posterior with four symmetrical modes. 

\textbf{Two moons:} This model has two parameters with a uniform prior. The simulator is non-linear, generating a posterior with both local and global (bimodal) structure \citep{greenberg2019automatic}.

\textbf{SIR:} Epidemiological model with two parameters and ten summary statistics \citep{kermack1927contribution}.

\subsection{Errors due to truncation}
\label{sec:appendix:epsilon_errors}
As described in Appendix Sec.~\ref{sec:appendix:proofconvergence}, the approximate posterior converges to the true posterior if the truncated proposal covers the support of the true posterior. The truncated support is defined as the high probability region that contains 1-$\epsilon$ of mass of the approximate posterior ($\text{HPR}_{\epsilon}$). For $\epsilon > 0$, the $\text{HPR}_{\epsilon}$ will likely not be a superset of the support of the true posterior and hence, there will be errors in posterior approximation. In this section, we discuss the effect of these errors on inference accuracy.

When the value of $\epsilon$ is chosen too large, the tails of the approximate posterior are excluded from the truncated proposal. In the following training round, the approximate posterior converges to a distribution that is correct, up to proportionality, within the $\text{HPR}_{\epsilon}$ of the previous approximate posterior, but that underestimates the tails of the posterior distribution. We demonstrate this behavior in Appendix Fig.~\ref{fig:fig4:supp7} for a linear Gaussian simulator, uniform prior, 50k simulations per round, and a neural spline flow with 20 bins as neural density estimator. After round 1, the approximate posterior closely matches the true posterior (Fig.~\ref{fig:fig4:supp7}a). When using a large $\epsilon$, e.g., $\epsilon = 0.1$, the proposal for the second round is narrower than the true posterior and, thus, the proposal obtained by pooling data from both rounds is not constant on the support of the true posterior (Fig.~\ref{fig:fig4:supp7}b, blue). This leads to the approximate posterior underestimating the tails of the true posterior (Fig.~\ref{fig:fig4:supp7}b, purple). When using a smaller $\epsilon$, e.g., $\epsilon = 0.01$, the errors induced by truncation become small and inference errors are mostly due to finite data and imperfect convergence of the neural network (Fig.~\ref{fig:fig4:supp7}c). We note that, throughout our study, we evaluated $\epsilon = \{10^{-3}, 10^{-4}, 10^{-5}\}$, i.e., values that are at least one order of magnitude smaller than $\epsilon = 0.01$.

\begin{figure}[ht!]
\ifdefined\NOFIGS
    [FIGURE SKIPPED]
    [Comment out line 3 of main.tex to include]
\else
    \includegraphics[width=\linewidth]{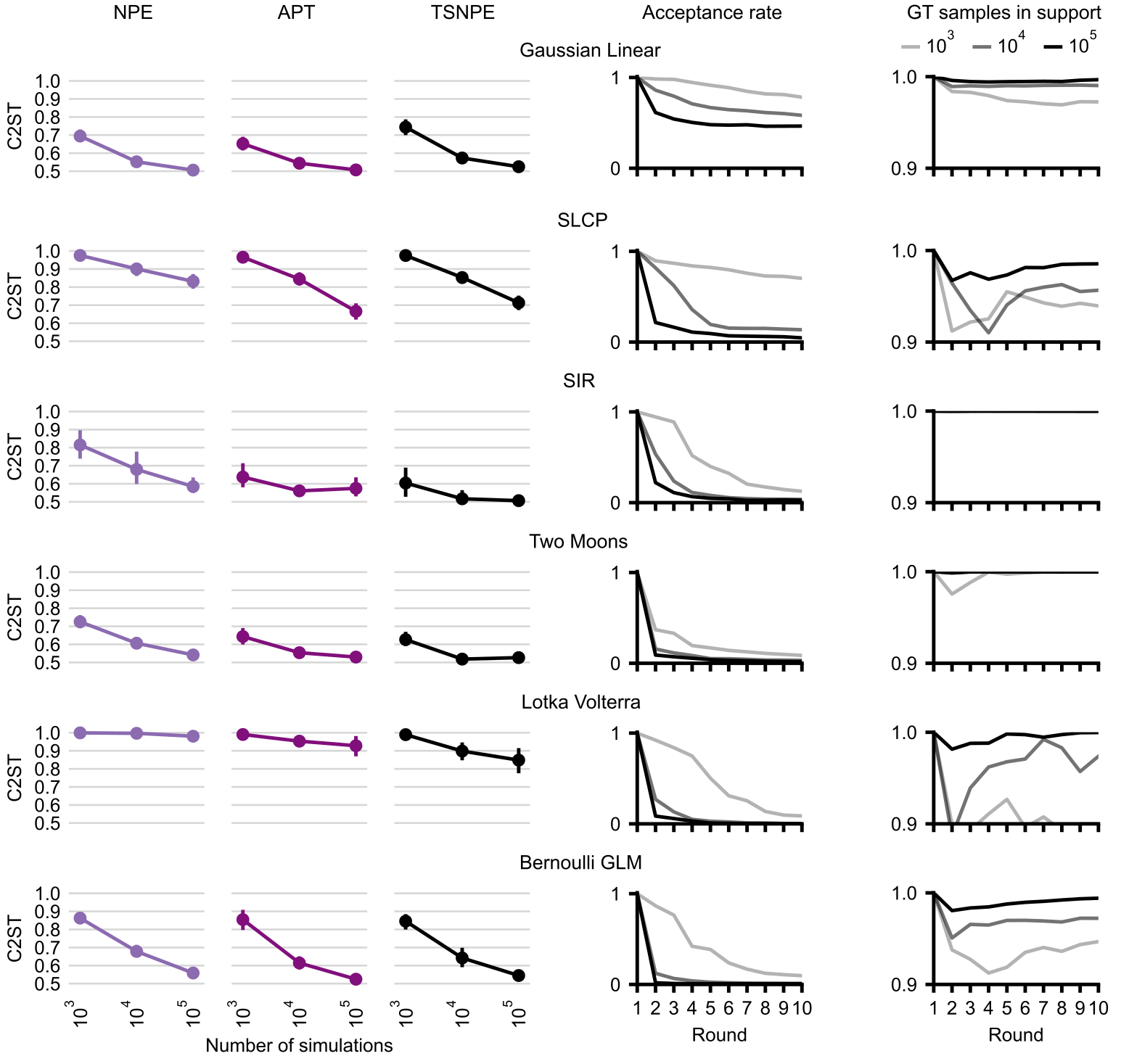}
\fi
\caption{
{\bf Benchmark results for a less conservative threshold.}
All hyperparameters are the same as in Fig.~\ref{fig:fig4}. The only difference is that we used $\epsilon = 10^{-3}$ (compared to $10^{-4}$ in Fig.~\ref{fig:fig4}).
}

\label{fig:fig4:supp1}
\end{figure}

\begin{figure}[ht!]
\ifdefined\NOFIGS
    [FIGURE SKIPPED]
    [Comment out line 3 of main.tex to include]
\else
    \includegraphics[width=\linewidth]{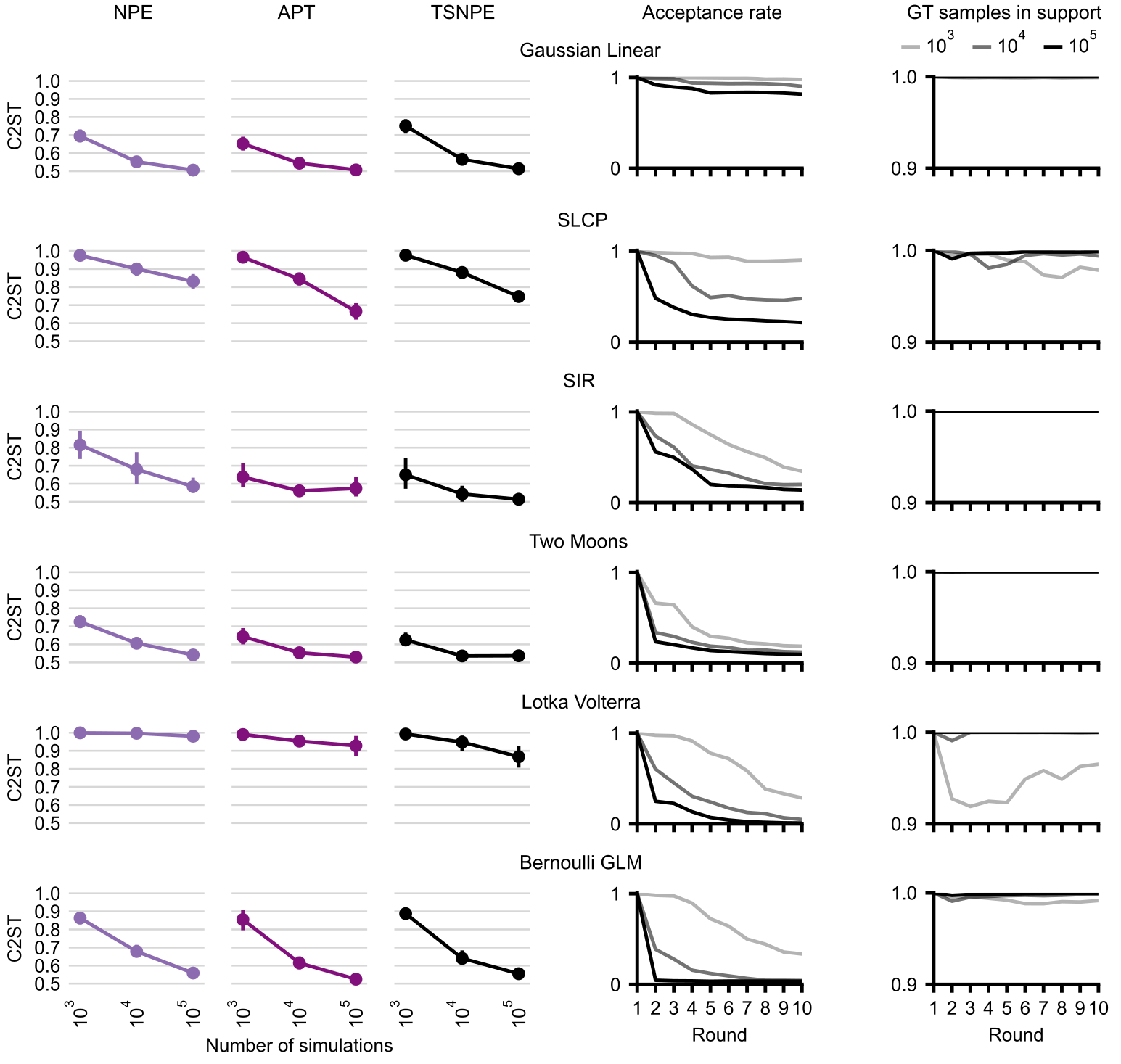}
\fi
\caption{
{\bf Benchmark results for a more conservative threshold.}
All hyperparameters are the same as in Fig.~\ref{fig:fig4}. The only difference is that we used $\epsilon = 10^{-5}$ (compared to $10^{-4}$ in Fig.~\ref{fig:fig4}).
}

\label{fig:fig4:supp2}
\end{figure}

Overall, this analysis demonstrates that the truncation performed by TSNPE can negatively impact inference quality in the tails of the posterior distribution. We, thus, do not recommend TSNPE in scenarios in which users are particularly interested in the tails of the posterior. In all our benchmark tasks, however, we did not find that the truncation negatively impacted inference quality as measured by C2ST accuracy (Fig.~\ref{fig:fig4}, Appendix Fig.~\ref{fig:fig4:supp1}, Fig.~\ref{fig:fig4:supp1}). This indicates that, for many (real-world) tasks, the errors due to truncation are outweighed by errors due to finite simulation budgets or imperfect convergence of neural network training.

\subsection{Computational cost of rejection sampling and SIR}
\label{sec:appendix:sampling_cost}
In this section, we quantify the computational costs of rejection sampling and sampling-importance resampling (SIR). The computational cost of both sampling methods comprises the computational cost of sampling and evaluating the approximate posterior. On an AMD Ryzen Threadripper 1920X 12-Core Processor, drawing (or evaluating) 100k samples takes approximately 10 seconds. On a GeForce RTX 2080 GPU, drawing (or evaluating) 100k samples takes approximately 0.17 seconds. Thus, in SIR (with an oversampling factor $K=1024$), one can draw (or evaluate) 100 samples from the truncated proposal in 0.17 seconds on a GPU (versus 10 seconds on a CPU). For most real-world simulators, this constitutes a small fraction of the compute time required to simulate the model: e.g., for the multicompartment model, a single simulation takes approximately 30 seconds and, thus, SIR sampling from the truncated support takes up only 0.012\% of the total compute time with a GPU. For rejection sampling, the time required to draw samples from the truncated support depends on the rejection rate. However, as long as the acceptance rate is above 0.001\%, the cost of rejection sampling is still small compared to the cost of running the simulator.

\subsection{Accuracy of SIR}
\label{sec:appendix:sampling_sir_accuracy}
\begin{figure}[ht!]
\ifdefined\NOFIGS
    [FIGURE SKIPPED]
    [Comment out line 3 of main.tex to include]
\else
    \includegraphics[width=\linewidth]{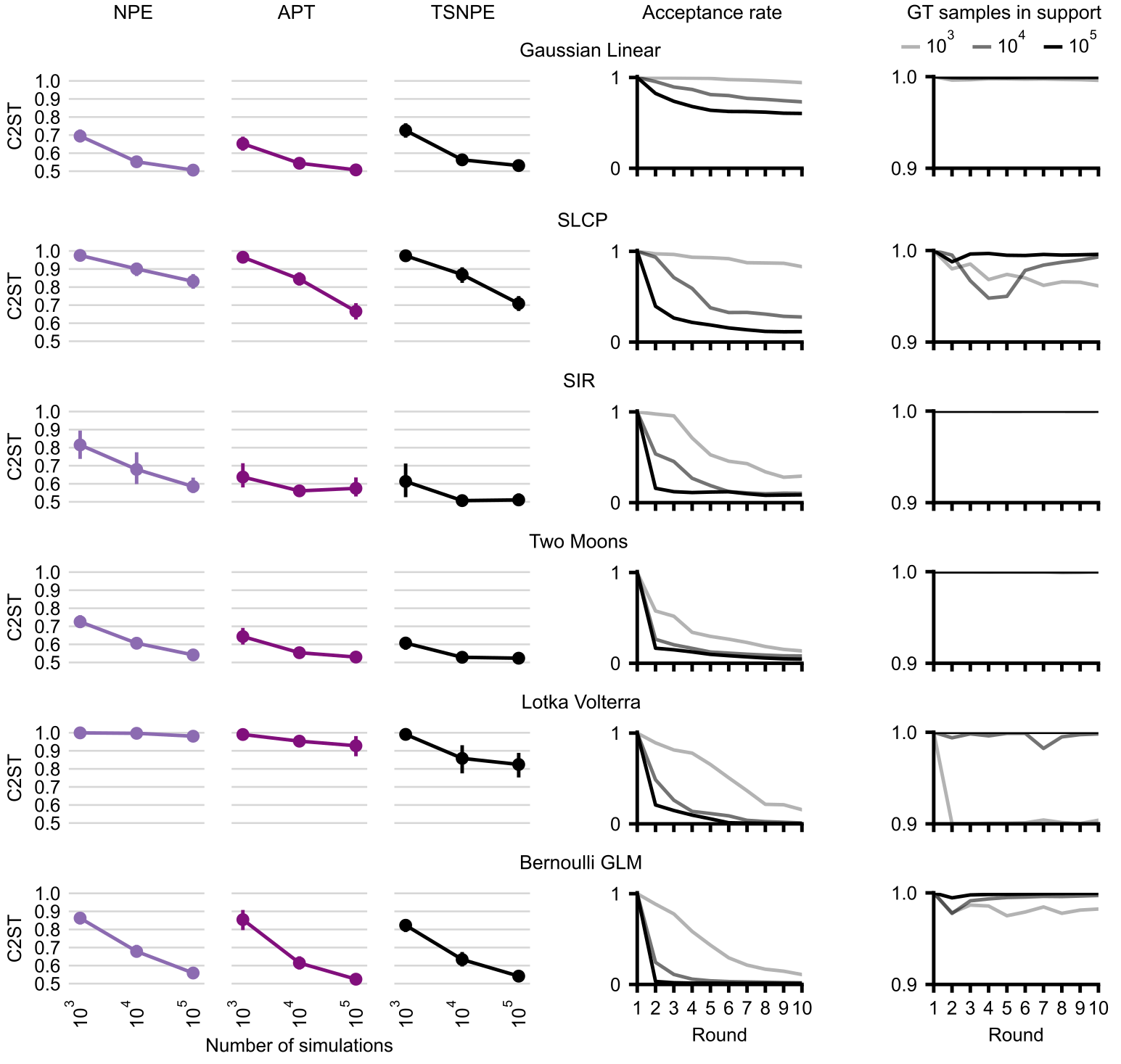}
\fi
\caption{
{\bf Benchmark results for sampling importance resampling (SIR).}
All hyperparameters are the same as in Fig.~\ref{fig:fig4} ($\epsilon = 10^{-4}$) but we use SIR to sample from the truncated proposal (instead of rejection sampling in Fig.~\ref{fig:fig4}).
}

\label{fig:fig4:supp3}
\end{figure}

Here, we investigate the (potential) error induced by using sampling-importance resampling (SIR). SIR is an approximate sampling technique and does not produce exact samples for finite $K$. This raises the question of how strongly the errors induced by SIR influence the results of TSNPE. In order to investigate this, we performed three analyses: 1) We ran all benchmarking tasks with SIR and $K=1024$ and compared the results to rejection sampling. Across all benchmark tasks, the performance of TSNPE with SIR matches the performance of TSNPE with rejection sampling (Appendix Fig.~\ref{fig:fig4:supp3}). 2) In a simple 1D toy model, we investigated how closely the samples produced by SIR match the samples produced by rejection sampling. As can be seen in Appendix Fig.~\ref{fig:fig4:supp4}, the distribution of SIR samples is quite different from rejection sampling for $K=16$ and $K=64$. However, for $K=1024$, the distribution of samples from SIR very closely matches the distribution of rejection samples. 3) Finally, we investigated the performance of SIR by inspecting the effective sample size (ESS), which we computed as 
\begin{equation*}
    \text{ESS} = 1 / \sum_i^K w_i^2
\end{equation*}
with $w_i$ being the normalized importance weights \citep{kong1992note, djuric2003particle, martino2017effective}. For $K=1024$, across all benchmark tasks, the ESS was on average 25.154 and was never below 2.547, i.e.~it was always significantly higher than 1 (the number of resampled samples). All of these results indicate that SIR is expected to be a useful and robust sampling method for TSNPE.

\subsection{Pyloric network model}
For the pyloric network model, we used the same prior, simulator, and summary statistics as previous work \citep{gonccalves2020training, deistler2021disparate, gloeckler2022variational}. The model has a total of $31$ parameters and $18$ summary statistics. We replaced invalid summary statistics with a value that is 2 standard deviations (of prior predictives) below the observation. The experimental data \citep{haddad10recordings} is also the same as used in these previous works.

\subsection{Multicompartment model of a single neuron}
\label{sec:appendix:l5pcmodel}

We performed Bayesian inference in a complex model of single-neuron dynamics. The model is the same as used in \citet{van2016bluepyopt}. We added observation noise with standard deviations taken from previously published measurements \citep{hay2011models}. The prior is a uniform distribution within the same bounds as previously used \citep{van2016bluepyopt}. The parameters are shown in Table \ref{table2}. The summary statistics are also the same as in \citet{van2016bluepyopt}. 

We replaced NaN values by the minimal value among prior samples minus two standard deviations of prior samples. Several summary statistics had heavy tailed distributions, which led to very high standard deviations. For these summary statistics, we picked the replacement value by hand. The final values are shown in Table \ref{table1}.

\begin{table}
\centering
\begin{tabular}{|c | c | c|} 
 \hline
 Index & Parameter & Ground truth \\ [0.5ex] 
 \hline\hline
 $\theta_1$ & gnats2\_tbar\_nats2\_t\_apical &  0.026145 \\
 \hline
 $\theta_2$ & gskv3\_1bar\_skv3\_1\_apical &  0.004226 \\
 \hline
 $\theta_3$ & gimbar\_im\_apical &  0.000143 \\
 \hline
 $\theta_4$ & gnata\_tbar\_nata\_t\_axonal &  3.137968 \\
 \hline
 $\theta_5$ & gk\_tstbar\_k\_tst\_axonal &  0.089259 \\
 \hline
 $\theta_6$ & gamma\_cadynamics\_e2\_axonal &  0.00291 \\
 \hline
 $\theta_7$ & gnap\_et2bar\_nap\_et2\_axonal &  0.006827 \\
 \hline
 $\theta_8$ & gsk\_e2bar\_sk\_e2\_axonal &  0.007104 \\
 \hline
 $\theta_9$ & gca\_hvabar\_ca\_hva\_axonal &  0.00099 \\
 \hline
 $\theta_{10}$ & gk\_pstbar\_k\_pst\_axonal &  0.973538 \\
 \hline
 $\theta_{11}$ & gskv3\_1bar\_skv3\_1\_axonal &  1.021945 \\
 \hline
 $\theta_{12}$ & decay\_cadynamics\_e2\_axonal &  287.19873 \\
 \hline
 $\theta_{13}$ & gca\_lvastbar\_ca\_lvast\_axonal &  0.008752 \\
 \hline
 $\theta_{14}$ & gamma\_cadynamics\_e2\_somatic &  0.000609 \\
 \hline
 $\theta_{15}$ & gskv3\_1bar\_skv3\_1\_somatic &  0.303472 \\
 \hline
 $\theta_{16}$ & gsk\_e2bar\_sk\_e2\_somatic &  0.008407 \\
 \hline
 $\theta_{17}$ & gca\_hvabar\_ca\_hva\_somatic &  0.000994 \\
 \hline
 $\theta_{18}$ & gnats2\_tbar\_nats2\_t\_somatic &  0.983955 \\
 \hline
 $\theta_{19}$ & decay\_cadynamics\_e2\_somatic &  210.48529 \\
 \hline
 $\theta_{20}$ & gca\_lvastbar\_ca\_lvast\_somatic &  0.000333 \\
 \hline
\end{tabular}
\caption{L5PC parameters.}
\label{table2}
\end{table}

\begin{table}
\centering
\begin{tabular}{|c | c | c|} 
 \hline
 Summary statistic & Observation & Replacement value\\ [0.5ex] 
 \hline\hline
 step1\_soma\_ahp\_depth\_abs & -62.1358 &  -110.974 \\
 \hline
step1\_soma\_ahp\_depth\_abs\_slow & -62.2882 & -151.52 \\ 
 \hline
step1\_soma\_ahp\_slow\_time & 0.140599 & -0.8473 \\ 
 \hline
step1\_soma\_ap\_height & 28.43591 & -33.959 \\ 
 \hline
step1\_soma\_ap\_width & 0.67857 & -2.132 \\ 
 \hline
step1\_soma\_isi\_cv & 0.03328 & -1.202 \\
 \hline
step1\_soma\_adaptation\_index2 & -0.0039499 & -0.3790 \\ 
 \hline
step1\_soma\_doublet\_isi & 67.00 & -1.699 \\  
 \hline
step1\_soma\_mean\_frequency & 7.106 & -52.343 \\ 
 \hline
step1\_soma\_time\_to\_first\_spike & 33.3000 & -719.1 \\ 
 \hline
 step2\_soma\_ahp\_depth\_abs & -60.6933 & -110.974 \\
 \hline
step2\_soma\_ahp\_depth\_abs\_slow & -60.8186 & -151.066 \\ 
 \hline
step2\_soma\_ahp\_slow\_time & 0.1496 & -0.8481 \\ 
 \hline
step2\_soma\_ap\_height & 26.5820 & -33.958 \\ 
 \hline
step2\_soma\_ap\_width & 0.67058 & -2.1747 \\ 
 \hline
step2\_soma\_isi\_cv & 0.03598 & -1.0649 \\
 \hline
step2\_soma\_adaptation\_index2 & -0.001467 & -0.47684 \\ 
 \hline
step2\_soma\_doublet\_isi & 44.600 & -1.6999 \\  
 \hline
step2\_soma\_mean\_frequency & 8.8444 & -67.842 \\ 
 \hline
step2\_soma\_time\_to\_first\_spike & 23.000 & -719.1 \\ 
 \hline
 step3\_soma\_ahp\_depth\_abs & -56.759 & -110.744 \\
 \hline
step3\_soma\_ahp\_depth\_abs\_slow & -55.903 & -149.126 \\ 
 \hline
step3\_soma\_ahp\_slow\_time & 0.2168 & -0.83586 \\ 
 \hline
step3\_soma\_ap\_height & 16.968 & -34.01 \\ 
 \hline
step3\_soma\_ap\_width & 0.5968 & -2.6932 \\ 
 \hline
step3\_soma\_isi\_cv & 0.09933 & -1.1164 \\
 \hline
step3\_soma\_adaptation\_index2 & 0.007206 & -0.5326 \\ 
 \hline
step3\_soma\_doublet\_isi & 21.100 & -1.699 \\  
 \hline
step3\_soma\_mean\_frequency & 16.086 & -125.30 \\ 
 \hline
step3\_soma\_time\_to\_first\_spike & 10.600 & -719.1 \\ 
 \hline
bap\_dend1\_ap\_amplitude\_from\_voltagebase & 53.267 & -60.701 \\
 \hline
bap\_dend2\_ap\_amplitude\_from\_voltagebase & 30.592 & -31.779 \\ 
 \hline
bap\_soma\_ap\_height & 37.519 & -33.95 \\  
 \hline
bap\_soma\_ap\_width & 0.800 & -1.583 \\ 
 \hline
bap\_soma\_spikecount & 1.0 & -0.8855 \\ 
 \hline
\end{tabular}
\caption{L5PC summary statistics.}
\label{table1}
\end{table}

\subsection{Choices of hyperparameters}
\label{sec:appendix:hyperparameters}
For the results on the benchmark tasks, we picked the same hyperparameters for TSNPE as those that were used in \citet{lueckmann2021benchmarking} for APT (called SNPE in \citet{lueckmann2021benchmarking}).

For both neuroscience tasks, we used a neural spline slow (NSF) as density estimator \citep{durkan2019neural}. The hyperparameters of the NSF are the defaults from the `sbi' package \citep{tejerocantero2020sbi}. On both of these tasks, we used $\epsilon = 10^{-3}$. All other hyperparameters are the defaults from the `sbi' package with one exception: We used a batchsize of $500$ (instead of the default value $50$).

For the pyloric network task, we ran APT with $2$ atoms. We reduced the number of atoms from the default value in the `sbi' package ($10$ atoms) because a larger number of atoms increased training time. When using $10$ atoms, the training time exceeded the simulation time of the model on the pyloric network task. With $2$ atoms, the training time of APT was comparable to the training time of TSNPE (albeit still a bit higher). We implemented the transformation of the parameter space with the pytorch method `biject\_to()' \citep{PyTorch2019Paszke}. For TSNPE, we initially sampled from the truncated proposal with rejection sampling. After the eighth round of training, the rejection rate became exceedingly high and we switched sampling importance resampling (SIR, with $K=1024$ see Sec.~\ref{sec:methodology:sampling}).

For the multicompartment model of single-neuron dynamics, we initially sampled parameters from the truncated proposal with rejection sampling and switched to SIR after the third round. For this task, in the first round of training, we used an ensemble of 10 neural networks. From the second round onward we used only a single neural network. For all other runs (toy example, benchmark, pyloric network), we did not use ensembles but always trained only a single network.

\begin{figure}[ht!]
\ifdefined\NOFIGS
    [FIGURE SKIPPED]
    [Comment out line 3 of main.tex to include]
\else
    \includegraphics[width=\linewidth]{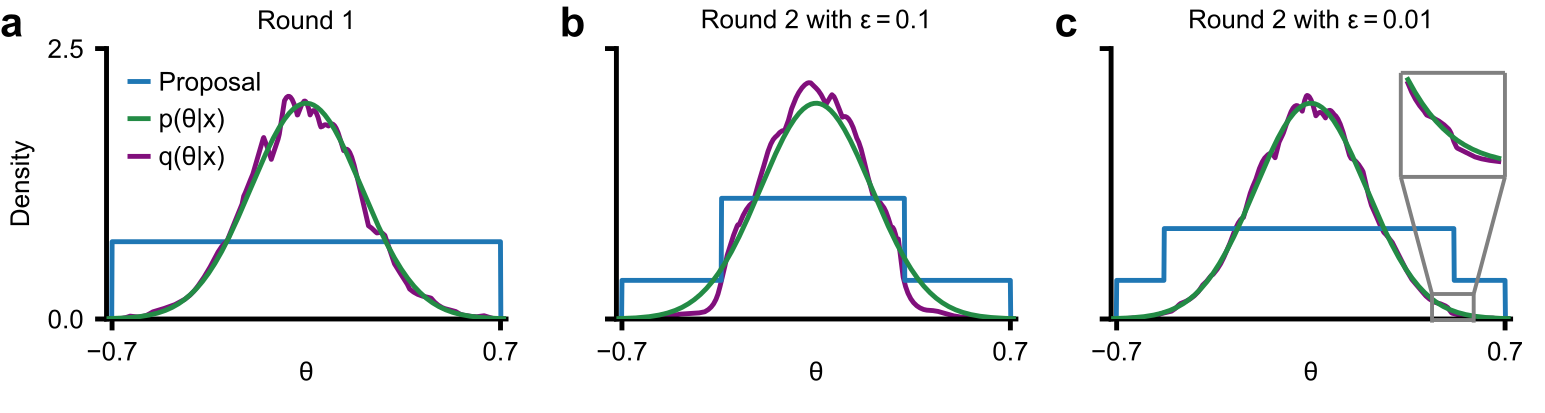}
\fi
\caption{
{\bf Errors induced by truncation.}
Inference in a linear Gaussian toy model with uniform prior.
(a) The approximate posterior after round 1 closely matches the true posterior.
(b) For a large truncation value  $\epsilon = 0.1$, the approximate posterior after round 2 systematically underestimates the tails of the true posterior distribution.
(c) For a smaller truncation value $\epsilon = 0.01$, the error induced by truncation is small.
}
\label{fig:fig4:supp7}
\end{figure}
\begin{figure}[ht!]
\ifdefined\NOFIGS
    [FIGURE SKIPPED]
    [Comment out line 3 of main.tex to include]
\else
    \includegraphics[width=\linewidth]{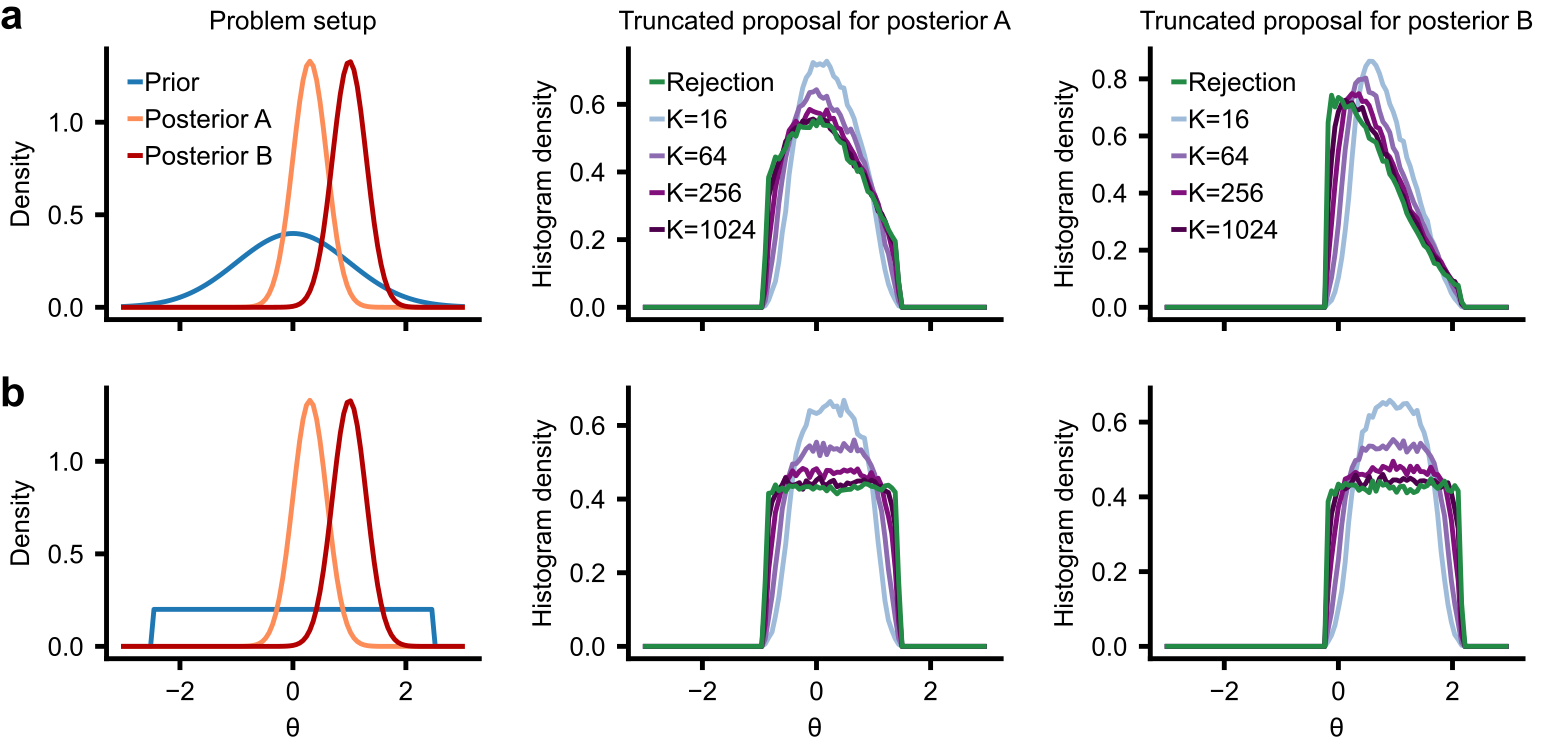}
\fi
\caption{
{\bf Comparison of truncated proposal between rejection and importance sampling.}
(a) Left: Gaussian prior as well as two posteriors. Middle: Density of 100k samples from the truncated proposal for posterior A. Green is rejection sampling, purple shaded colors are SIR with different oversampling factors $K$. Right: Same as middle, but for posterior B.
(b) Same as panel a, but for a uniform prior.
}

\label{fig:fig4:supp4}
\end{figure}
\begin{figure}[ht!]
\ifdefined\NOFIGS
    [FIGURE SKIPPED]
    [Comment out line 3 of main.tex to include]
\else
    \includegraphics[width=\linewidth]{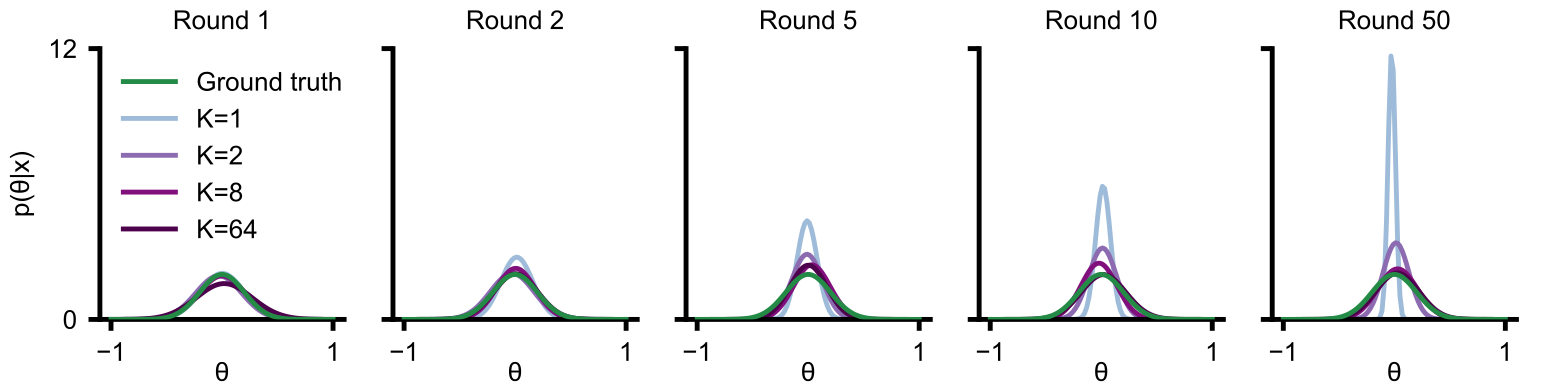}
\fi
\caption{
{\bf Poor hyperparameter choices can lead SIR to diverge.}
We applied TSNPE to a model with uniform prior (in [-1, 1]) and a linear Gaussian simulator. In each round, we ran 500 simulations, trained only on data from the most recent round, and used a Gaussian approximate posterior. We sampled from the truncated proposal with SIR (with different oversampling factors $K$). As more rounds are being run, the TSNPE approximate posterior can become too narrow for small $K$. Larger values of $K$ are robust across 50 rounds.
}
\label{fig:fig4:supp6}
\end{figure}
\begin{figure}[ht!]
\ifdefined\NOFIGS
    [FIGURE SKIPPED]
    [Comment out line 3 of main.tex to include]
\else
    \includegraphics[width=\linewidth]{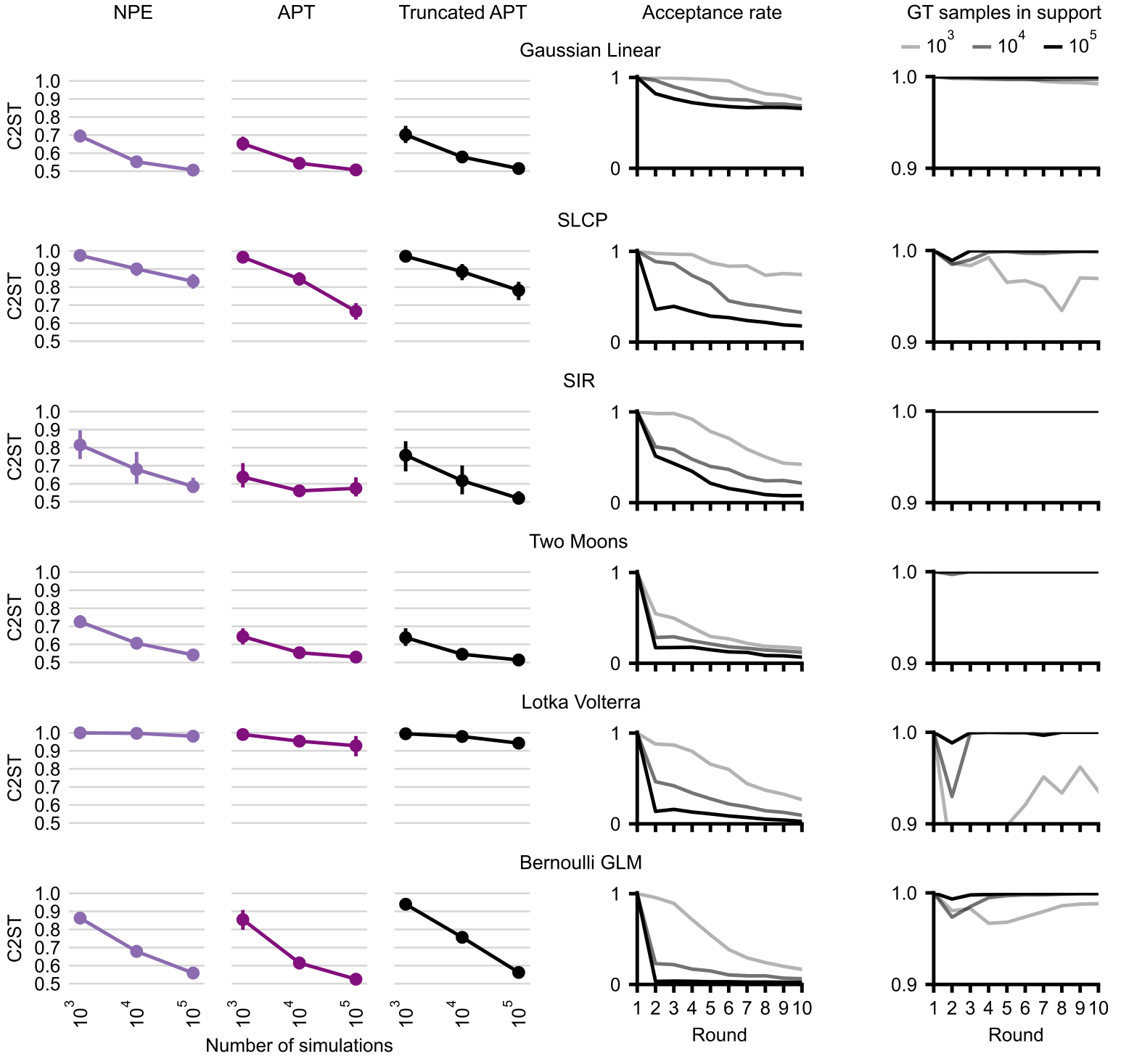}
\fi
\caption{
{\bf Comparison between APT and truncated APT.}
In order to investigate the effect of the truncated proposals on inference quality, we evaluated two versions of APT: In one scenario (middle column), we draw proposal samples from the previously estimated posterior, whereas in the second scenario (right column), we draw proposal samples from the truncated prior (with $\epsilon=10^{-4}$). In both versions, we used the atomic loss function proposed in \citet{greenberg2019automatic}. Therefore, all differences stem from the different proposals. We also compare to standard NPE (trained with maximum-likelihood loss, left column). The two columns on the right are the same as in Fig.~\ref{fig:fig4}, evaluated for truncated APT.
}

\label{fig:fig4:supp5}
\end{figure}
\begin{figure}[ht!]
\ifdefined\NOFIGS
    [FIGURE SKIPPED]
    [Comment out line 3 of main.tex to include]
\else
    \includegraphics[width=\linewidth]{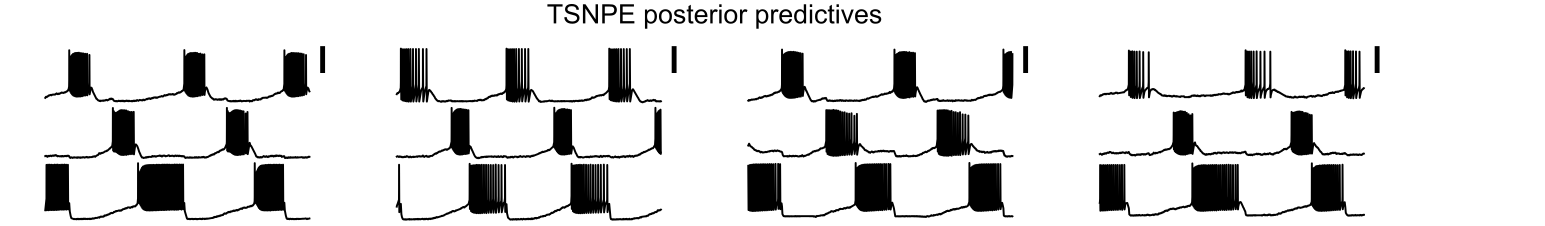}
\fi
\caption{
{\bf Four posterior predictives of TSNPE applied to the pyloric network model.
}
The generated activity closely matches the summary statistics of the experimental data (Fig.~\ref{fig:fig7}a).
}

\label{fig:fig7:supp2}
\end{figure}
\begin{figure}[ht!]
\ifdefined\NOFIGS
    [FIGURE SKIPPED]
    [Comment out line 3 of main.tex to include]
\else
    \includegraphics[width=\linewidth]{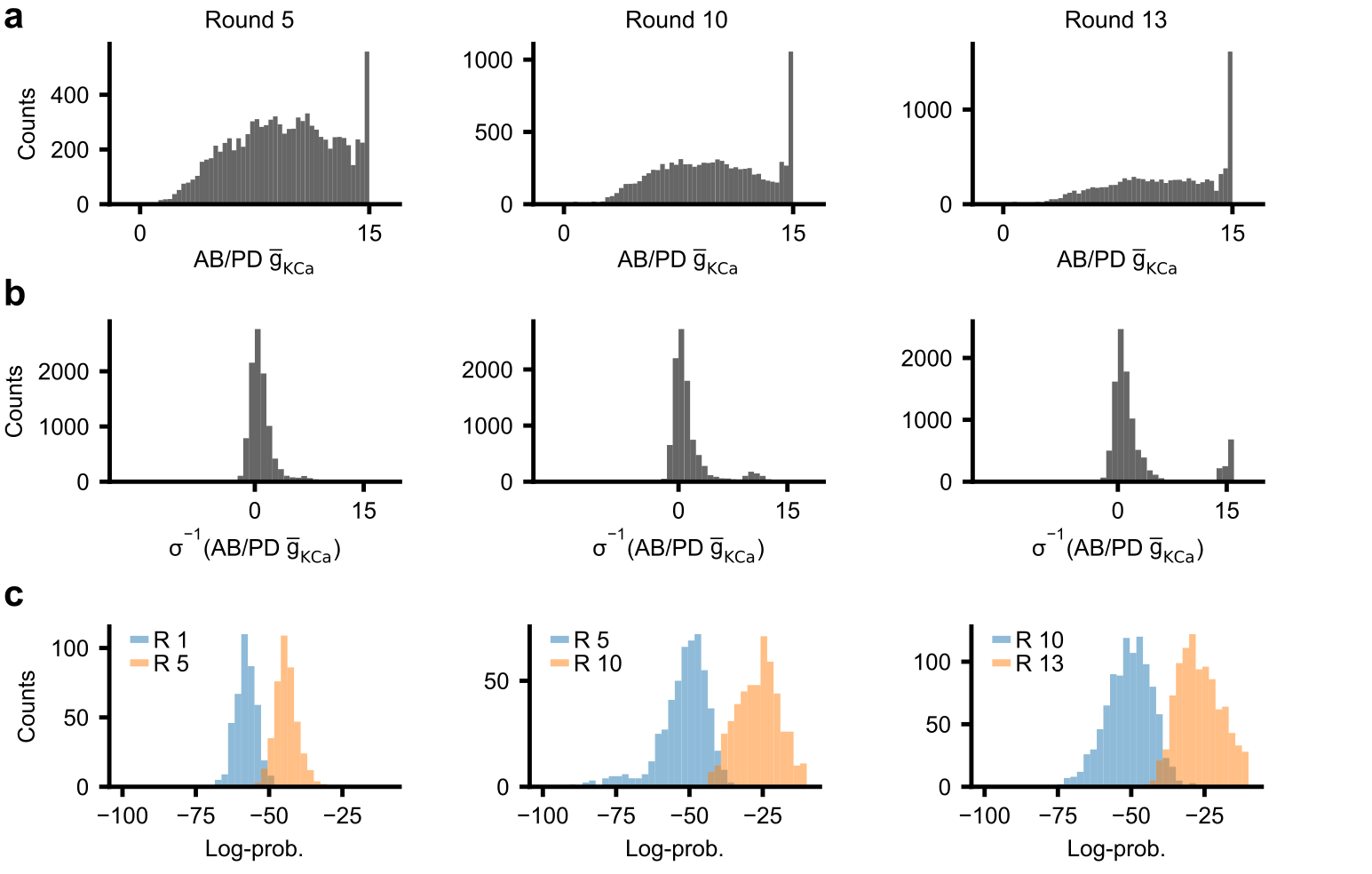}
\fi
\caption{
{\bf Further explanation of issues when the parameter space is constrained.
}
(a) Marginal distribution of $\overline{g}_{\text{KCa}}$ of the AB/PD neuron when running APT for 5, 10,  and 13 rounds.
(b) Marginal distribution of this parameter, but when transformed into unconstrained space. This is the training data that the normalizing flow is `effectively' seeing.
(c) We evaluated those samples that were at the very right bounds of the marginal distribution under the current posterior (blue) as well as under the posterior from approximately 5 rounds before (orange). The samples have consistently lower log-probability under the previous posterior, which shows that the location of the leaking mass is moving.
}

\label{fig:fig7:supp4}
\end{figure}
\begin{figure}[ht!]
\ifdefined\NOFIGS
    [FIGURE SKIPPED]
    [Comment out line 3 of main.tex to include]
\else
    \includegraphics[width=\linewidth]{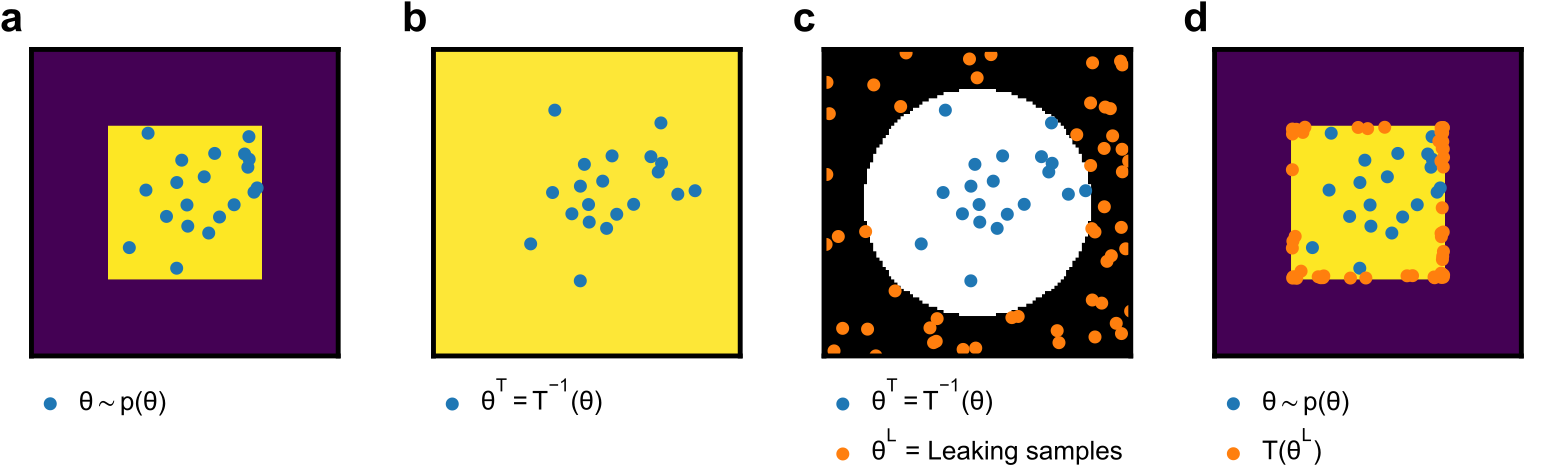}
\fi
\caption{
{\bf Illustration of `leakage' when prior is transformed to unconstrained space.
}
Note that these results are not based on an actual run of APT, but are purely for illustrative purposes.
(a) The prior is a two-dimensional distribution on a constrained space (yellow region). Samples from the prior are in blue.
(b) When the prior is transformed with an inverse sigmoid, its support becomes unconstrained.
(c) APT will fit the posterior within the region of the training data (white region, blue samples). Outside of the training data, the trained density estimator can put arbitrary mass without affecting the loss (black region). Samples from this region are in orange.
(d) When these `leaking' samples (obtained with APT) are transformed back into constrained space, the `leaking' mass ends up on the bounds of the prior.
}

\label{fig:fig7:supp5}
\end{figure}
\begin{figure}[ht!]
\ifdefined\NOFIGS
    [FIGURE SKIPPED]
    [Comment out line 3 of main.tex to include]
\else
    \includegraphics[width=\linewidth]{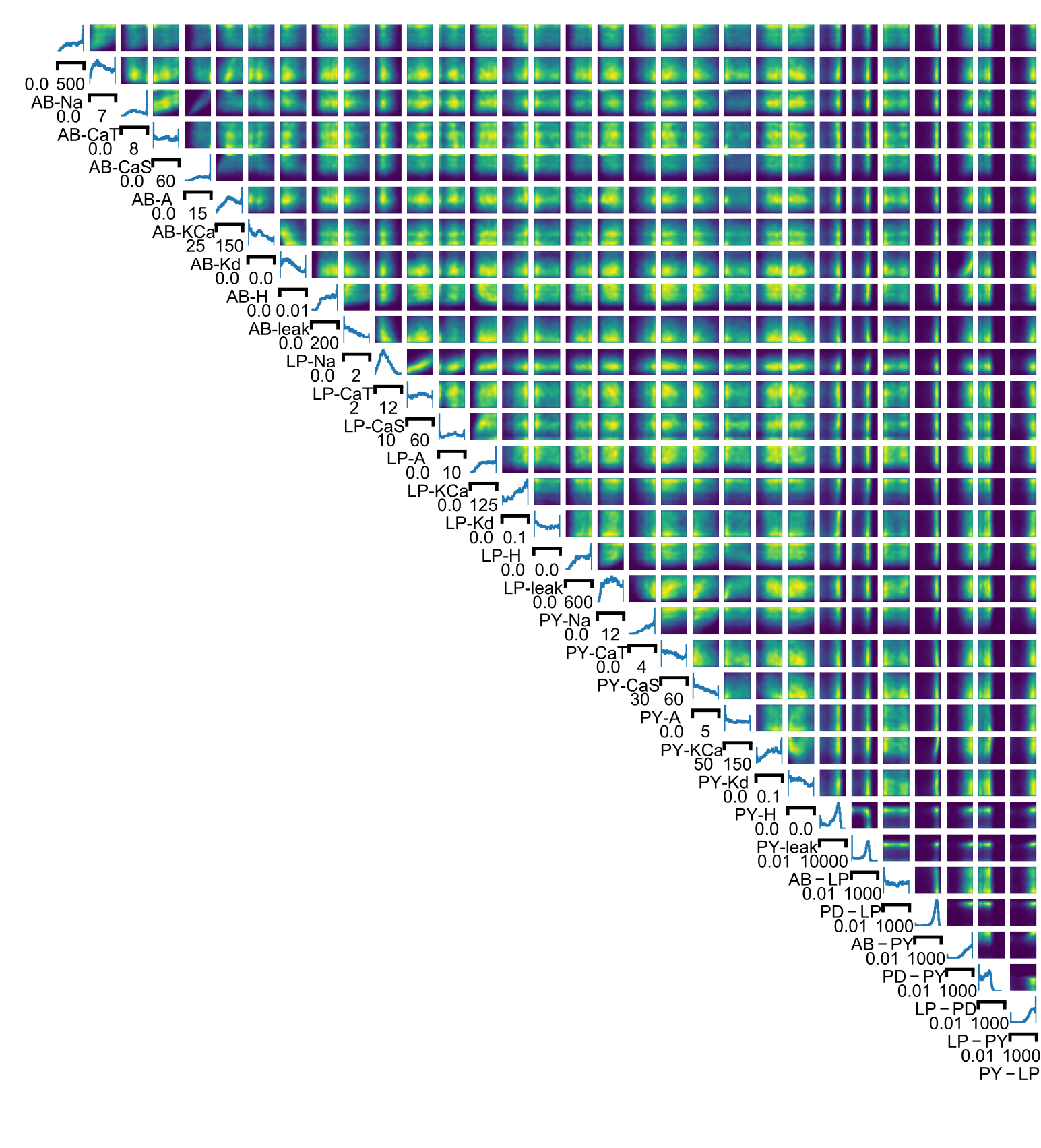}
\fi
\caption{
{\bf Posterior distribution inferred by APT for the pyloric network model.
}
}

\label{fig:fig7:supp6}
\end{figure}
\begin{figure}[ht!]
\ifdefined\NOFIGS
    [FIGURE SKIPPED]
    [Comment out line 3 of main.tex to include]
\else
    \includegraphics[width=\linewidth]{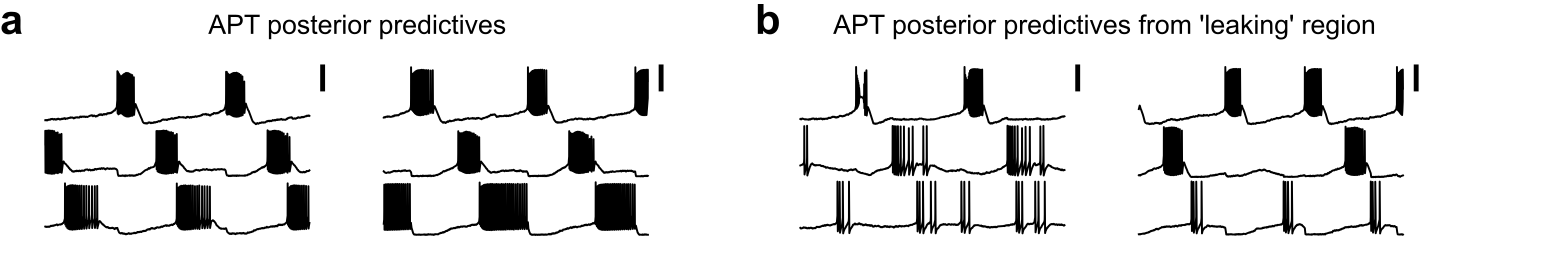}
\fi
\caption{
{\bf Failure of APT when transformed to constrained space.
}
(a) Activity generated by two parameter sets sampled from the posterior distribution obtained with APT. We ensured that the shown parameter sets are not sampled from the very bounds of posterior distribution (i.e., that they were not sampled from the peak shown in Fig.~\ref{fig:fig7}c).
(b) Activity generated by two parameter sets sampled from the posterior distribution obtained with APT. We specifically selected parameter sets whose value of $\overline{g}_{\text{KCa}}$ in the AB-PD neuron was above 14.999 (i.e., those samples which are in the peak shown in Fig.~\ref{fig:fig7}c).
}

\label{fig:fig7:supp1}
\end{figure}
\begin{figure}[ht!]
\ifdefined\NOFIGS
    [FIGURE SKIPPED]
    [Comment out line 3 of main.tex to include]
\else
    \includegraphics[width=\linewidth]{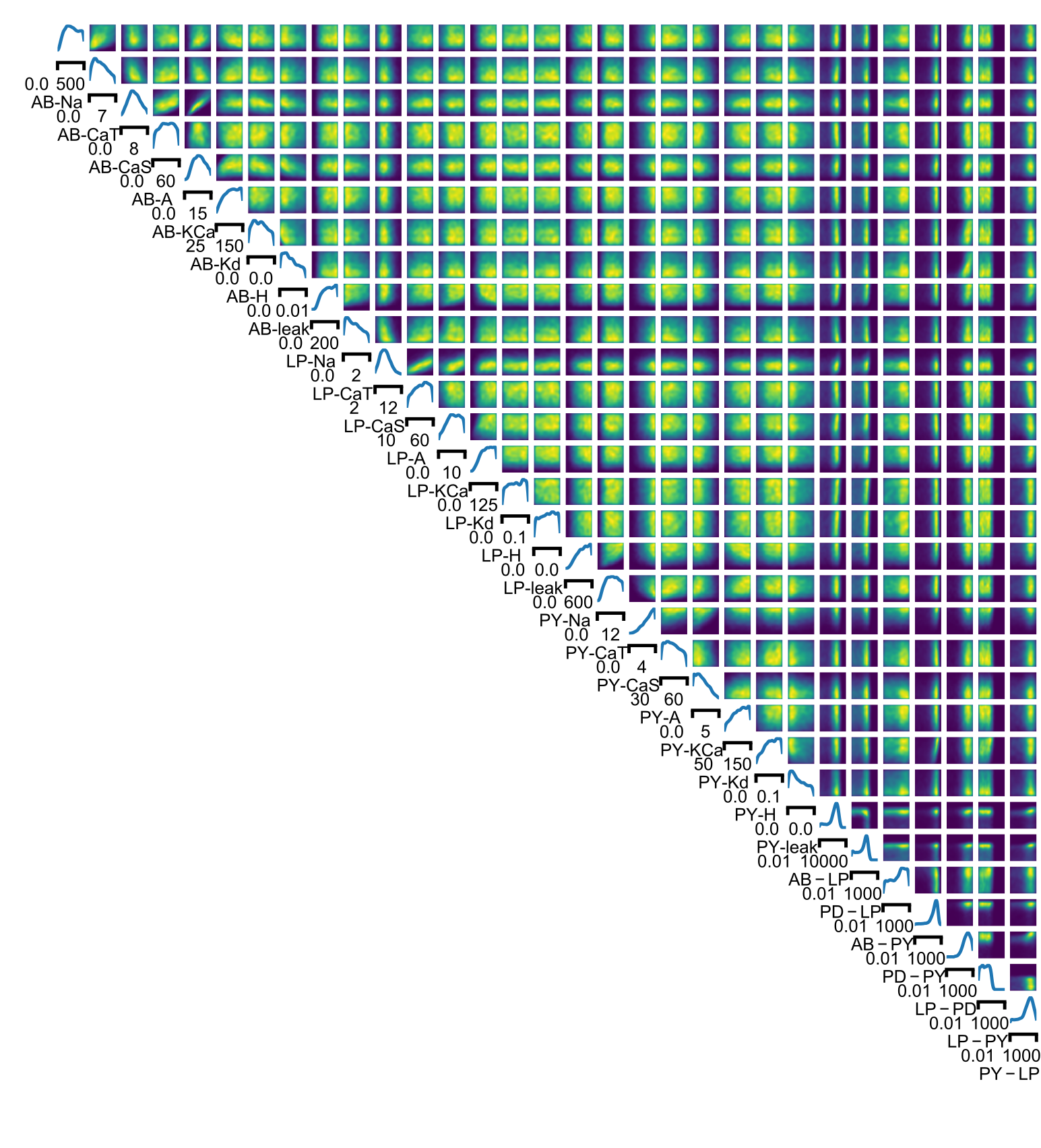}
\fi
\caption{
{\bf Posterior distribution inferred by TSNPE for the pyloric network model.
}
}

\label{fig:fig7:supp3}
\end{figure}
\begin{figure}[ht!]
\ifdefined\NOFIGS
    [FIGURE SKIPPED]
    [Comment out line 3 of main.tex to include]
\else
    \includegraphics[width=\linewidth]{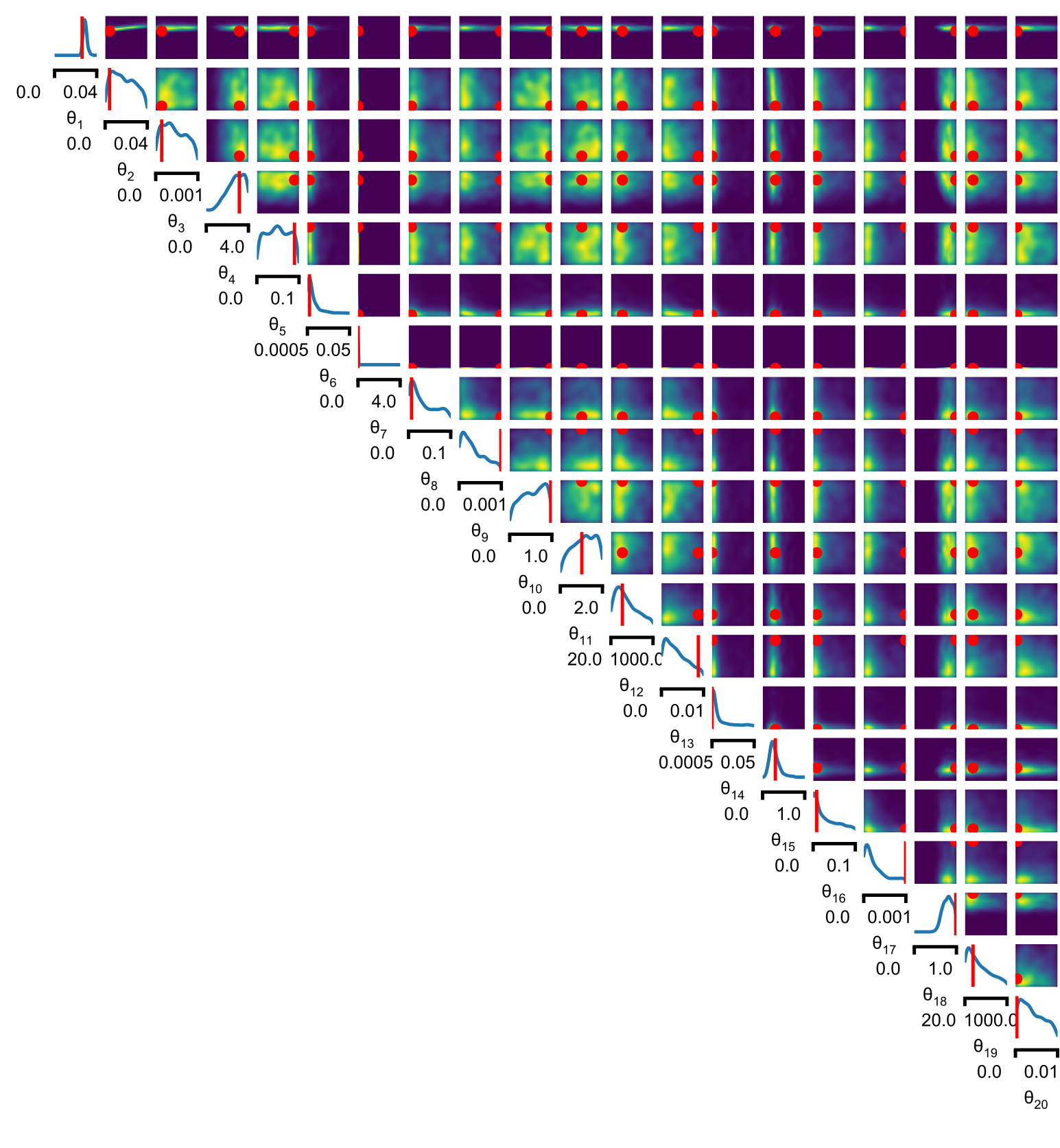}
\fi
\caption{
{\bf Full posterior distribution over biophysical parameters for the layer 5 pyramidal cell.
}
Parameter names can be found in Appendix Sec.~\ref{sec:appendix:l5pcmodel}. Red dots are the ground-truth parameters.
}

\label{fig:fig5:supp1}
\end{figure}
\begin{figure}[ht!]
\ifdefined\NOFIGS
    [FIGURE SKIPPED]
    [Comment out line 3 of main.tex to include]
\else
    \includegraphics[width=\linewidth]{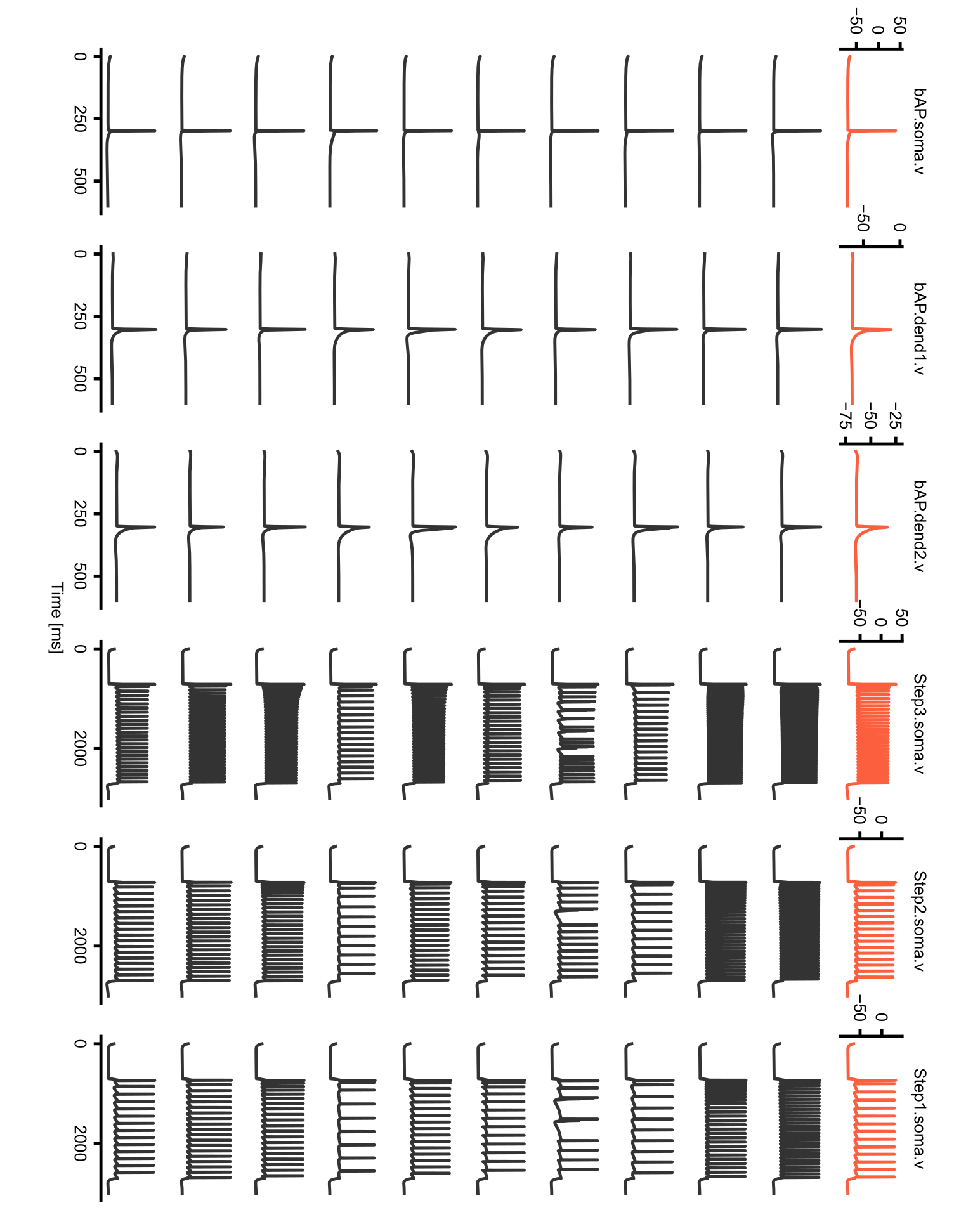}
\fi
\caption{
{\bf Observed data (orange, top row) and ten posterior predictives obtained with TSNPE (black).
}
}

\label{fig:fig5:supp2}
\end{figure}
\begin{figure}[ht!]
\ifdefined\NOFIGS
    [FIGURE SKIPPED]
    [Comment out line 3 of main.tex to include]
\else
    \includegraphics[width=\linewidth]{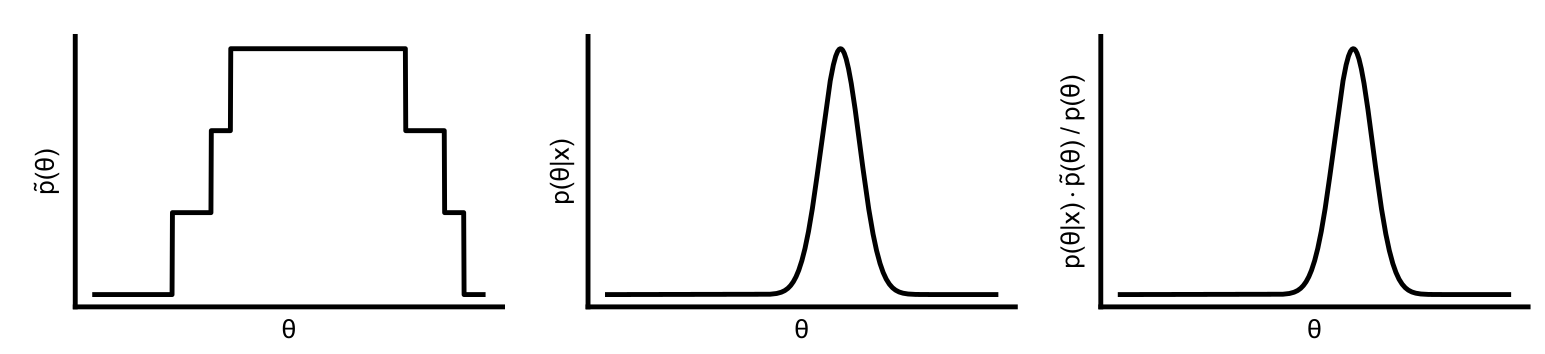}
\fi
\caption{
{\bf Truncated proposals when pooling data from multiple rounds.
}
Assume a uniform prior. Left: Proposal that is the average of the proposal over three rounds.
Middle: true posterior.
Right: the approximate posterior converges to $p(\btheta | \bx_o) \frac{\tilde{p}(\btheta)}{p(\btheta)}$ \citep{papamakarios2016fast, lueckmann2017flexible, greenberg2019automatic}. This matches the true posterior even though the proposal is not constant \emph{outside} of the $\text{HPR}_{\epsilon}$ of the true posterior.
}

\label{fig:fig6:supp1}
\end{figure}

\end{document}